\documentclass[lettersize,journal]{IEEEtran}
\usepackage{amsmath,amsfonts}
\usepackage{algorithmic}
\usepackage{algorithm}
\usepackage{array}
\usepackage[caption=false,font=normalsize,labelfont=sf,textfont=sf]{subfig}
\usepackage{textcomp}
\usepackage{stfloats}
\usepackage{url}
\usepackage{verbatim}
\usepackage{graphicx}
\usepackage{cite}
\usepackage{xcolor}
\usepackage{multirow}
\usepackage{booktabs}
\begin{document}

\title{Hierarchical Semantic-Visual Fusion of Visible and Near-infrared Images for Long-range Haze Removal}

\author{Yi Li, Xiaoxiong Wang, Jiawei Wang, Yi Chang*, Kai Cao, Luxin Yan
\thanks{This work was supported in part by the National Natural Science Foundation of China under Grant No. 62371203, and was supported in part by the Open Fund of the State Key Laboratory of Dynamic Optical Imaging and Measurement \textit{(Corresponding author: Yi Chang.)}}
\thanks{Y. Li, X. Wang, J. Wang, Y. Chang, K. Cao, and L. Yan are with National Key Laboratory of Multispectral Information Intelligent Processing Technology, School of Artificial Intelligence and Automation, Huazhong University of Science and Technology, Wuhan, China (E-mail: li\_yi@hust.edu.cn, xiaoxiongwang01@gmail.com, wjw328546818@163.com, yichang@hust.edu.cn, carl\_cao@hust.edu.cn, and yanluxin@hust.edu.cn). }
\thanks{Y. Chang is also with State Key Laboratory of Dynamic Optical Imaging and Measurement, Changchun, China. }
\thanks{The computation is completed in the HPC Platform of Huazhong University of Science and Technology.}
}

\markboth{IEEE TRANSACTIONS ON MULTIMEDIA}%
{Shell \MakeLowercase{\textit{et al.}}: A Sample Article Using IEEEtran.cls for IEEE Journals}

\maketitle

\begin{abstract}

While image dehazing has advanced substantially in the past decade, most efforts have focused on short-range scenarios, leaving long-range haze removal under-explored. As distance increases, intensified scattering leads to severe haze and signal loss, making it impractical to recover distant details solely from visible images. Near-infrared, with superior fog penetration, offers critical complementary cues through multimodal fusion. However, existing methods focus on content integration while often neglecting haze embedded in visible images, leading to results with residual haze. In this work, we argue that the infrared and visible modalities not only provide complementary low-level visual features, but also share high-level semantic consistency. Motivated by this, we propose a Hierarchical Semantic-Visual Fusion (HSVF) framework, comprising a semantic stream to reconstruct haze-free scenes and a visual stream to incorporate structural details from the near-infrared modality. The semantic stream first acquires haze-robust semantic prediction by aligning modality-invariant intrinsic representations. Then the shared semantics act as strong priors to restore clear and high-contrast distant scenes under severe haze degradation. In parallel, the visual stream focuses on recovering lost structural details from near-infrared by fusing complementary cues from both visible and near-infrared images. Through the cooperation of dual streams, HSVF produces results that exhibit both high-contrast scenes and rich texture details. Moreover, we introduce a novel pixel-aligned visible-infrared haze dataset with semantic labels to facilitate benchmarking. Extensive experiments demonstrate the superiority of our method over state-of-the-art approaches in real-world long-range haze removal.

\end{abstract}


\section{Introduction}
\IEEEPARstart{H}{aze}, especially long-range haze, has been shown to not only degrade visual quality substantially, but also impair numerous high-level vision tasks \cite{jia2012two, guo2021heterogeneous, wei2018deep, bijelic2020seeing, qian2021robust, dai2020curriculum, ma2022both, he2025exploring, cheng2025semantic, cheng2022hybrid}. Fundamentally, haze degradation suppresses scene-relevant signals and significantly reduces image contrast and signal-to-noise ratio, thereby weakening discriminative image features. Hence, dehazing aims to counteract these effects by recovering visually clear and structurally faithful scenes. Haze degradation is commonly modeled by the atmospheric scattering model \cite{narasimhan2002vision, li2017haze}:
\begin{equation}\
	\begin{aligned}
		\resizebox{!}{\height}{$I(x) = J(x)t(x)+A(1 - t(x)),$}
		\label{eq1}
	\end{aligned}
\end{equation}
where $I(x)$ and $J(x)$ are the hazy and latent haze-free images respectively, and $x$ represents pixel coordinates. $A$ is the global atmospheric light. $t(x)=e^{-\beta d(x)}$ is the transmission map, where $\beta$ and $d(x)$ represent the scattering coefficient and scene depth. As $\beta$ and $d(x)$ increase, the degradation becomes more severe, potentially resulting in the loss of scene information.

\begin{figure}[t]
	\centering
	\includegraphics[width=\linewidth]{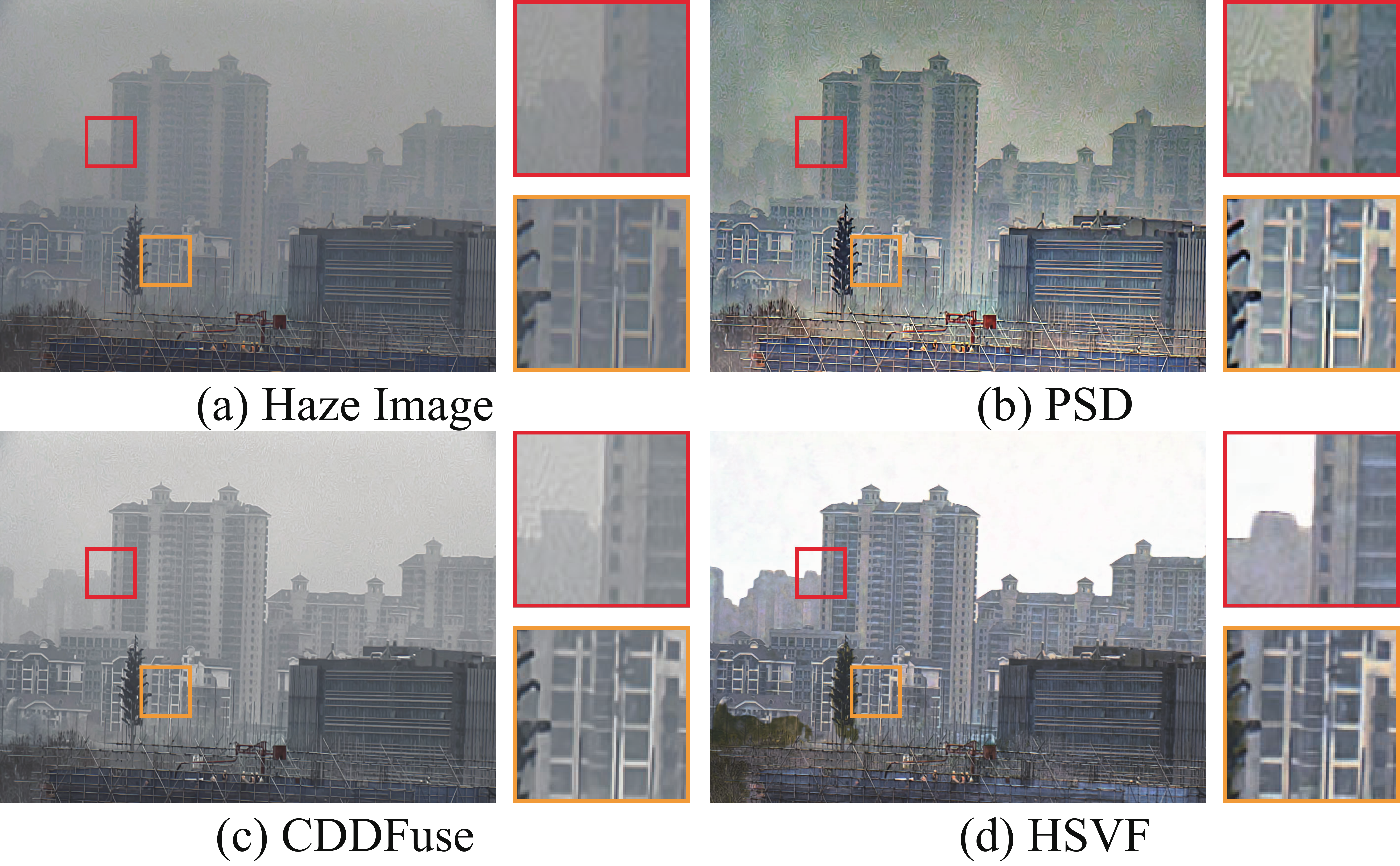}
	\caption{Visual comparison on haze removal with dehazing method PSD \cite{chen2021psd} and fusion method CDDFuse \cite{zhao2023cddfuse}. PSD handles short-range haze effectively (\textcolor{orange}{orange}), but struggles in distant regions (\textcolor{red}{red}) due to severe signal attenuation. CDDFuse enhances texture but leaves haze residuals. In contrast, our HSVF restores both global contrast and fine details, particularly in distant areas.}
	\label{fig_topfig}
\end{figure}

To mitigate the haze effect, image dehazing has been extensively studied over the past decades. Supervised dehazing methods, empowered by the strong representation capabilities of CNNs \cite{ren2016single, li2017aod, wu2021contrastive, liu2022towards, li2022single, zheng2023curricular, cheng2024continual} and transformers \cite{gao2022novel, song2023vision, qiu2023mb, liu2023data, liu2023visual}, have achieved substantial advancements. However, these methods rely heavily on synthetic paired data for supervised training, which limits their generalization to real-world haze conditions due to domain discrepancies. To address this limitation, researchers have explored semi-supervised \cite{li2019semi, shao2020domain, liu2021synthetic, chen2021psd, yang2022self} and unsupervised \cite{li2020zero, li2021you, zhao2021refinednet, li2022usid, ding2023u, wang2024ucl, liang2025image} strategies that aim to enhance generalization to real-world haze by bridging the domain gap via domain adaptation. Nevertheless, such methods mainly perform direct image-level translation across domains, yet the domain gap remains substantial, making knowledge transfer difficult. More fundamentally, these methods perform poorly on long-range haze, where distant textures are attenuated or even lost at the source due to severe scattering, as shown in Fig. \ref{fig_topfig}(b).

Owing to its longer wavelength, the near-infrared spectrum exhibits stronger resilience to haze-induced scattering than the visible spectrum, resulting in improved preservation of fine scene details such as textures and edges. Motivated by this property, another line of research focuses on utilizing near-infrared information to recover scene details that are degraded or lost in visible images under hazy conditions \cite{lex2009color, ahn2011haze, ma2022swinfusion, zhao2022efficient, liu2022learning, zhu2023near, zhao2023cddfuse}. These methods typically reformulate haze removal as a visible and near-infrared image fusion task, aiming to generate results with sharper edges and richer textures. Despite improving structure visibility, these methods often neglect the residual haze present in visible images, resulting in haze remnants and reduced contrast, as shown in Fig. \ref{fig_topfig}(c). 

In contrast to previous methods, this work is rooted in a key argument: visible and near-infrared modalities \emph{not only exhibit complementary low-level structures, but also share high-level semantic consistency}. Motivated by this observation, we propose a novel \textbf{H}ierarchical \textbf{S}emantic-\textbf{V}isual \textbf{F}usion (HSVF) framework that jointly leverages structural and semantic cues across modalities for effective long-range haze removal.

The proposed HSVF consists of two complementary streams: a semantic stream and a visual stream. The semantic stream recovers haze-free scenes guided by modality-invariant semantics as reliable category-level priors. The visual stream restores fine structures by integrating complementary visual cues from both modalities. Specifically, the semantic stream consists of two modules. The intrinsic semantic alignment module aligns high-level semantics across modalities to derive modality-invariant representations. Then the cross-domain semantic reconstruction module leverages the aligned semantics as a reliable intermediate bridge between the clear and hazy domains. This design enables effective knowledge transfer and facilitates category-aware haze-free reconstruction. In contrast, the visual stream focuses on low-level structure preservation by integrating visual cues from both modalities through a cross-modal visual fusion module. This module employs joint self- and cross-attention mechanisms to adaptively extract and fuse features across modalities. This hierarchical design enhances global scene contrast and restores fine details simultaneously, effectively addressing the challenging long-range haze, as shown in Fig. \ref{fig_topfig}(d).
	
Furthermore, we capture and construct a novel real-world dataset named VNHD, which consists of 1519 and 1442 pixel-aligned visible and near-infrared image pairs under hazy and clear conditions, along with semantic labels for further research. The main contributions are summarized  as follows:

\begin{itemize}
	\item We focus on the challenging and under-explored task of long-range haze removal, where intensified scattering leads to severe signal attenuation and irreversible scene information loss, making it impractical to recover distant scene details solely from degraded visible images.
	\item We propose a hierarchical semantic-visual fusion framework, which comprises a semantic stream for reconstructing clear high-contrast scenes and a visual stream for recovering fine-grained structures. By jointly leveraging high-level semantics and low-level visual cues, the proposed HSVF effectively enhances both scene clarity and texture richness for long-range haze images.
	\item We construct a novel real-world multimodal dataset VNHD, with 1519 and 1442 aligned visible-near-infrared image pairs and semantic labels under hazy and clear conditions, which serves as a valuable benchmark to facilitate future works. Extensive experiments on real-world long-range haze removal show that our HSVF consistently outperforms state-of-the-art approaches.
\end{itemize}

\section{Related Works}
\label{sec:related}
\noindent\textbf{Long-range haze removal.}
Image dehazing has been extensively studied over the past decade. Although early prior-based methods offered interpretable solutions from handcrafted cues \cite{tan2008visibility, he2010single, zhu2015fast, berman2016non, berman2017air, 8930996, 9537303, berman2018single}, they have been largely outperformed by learning-based approaches with stronger representational capacity \cite{cai2016dehazenet, ren2016single, li2017aod, li2018single, liu2019griddehazenet, chen2019gated, qu2019enhanced, liu2019learning, dong2020multi, dong2020physics, wu2021contrastive, liu2022towards, guo2022image, gao2022novel, jin2022structure, zheng2023curricular, song2023vision, qiu2023mb, liu2023data, liu2023visual, 8734728, 8792133, 8851251, 9316931, 9473023, 9726872, 9745359, 9794593, li2022single, cheng2024continual, liang2025image}. While these approaches have yielded notable results in short-range scenarios, long-range haze removal remains a significantly under-explored problem. Some recent works aim to enhance distant visibility \cite{fade2014long, zhu2018haze, liu2023lrinet, wang2024depth, liang2025image}. For example, Zhu \emph{et al.} combined the dark channel prior and brightness model to recover hazy skies \cite{zhu2018haze}, while Wang \emph{et al.} introduced a depth-guided network trained on synthetic data \cite{wang2024depth} for UAV-captured long-range scenarios. Despite significant progress, these methods still struggle with real-world long-range haze, where scene structures are often lost at acquisition. Instead of relying solely on degraded visible images, we incorporate near-infrared as an auxiliary modality to recover both low-level structures and high-level semantics. To fully exploit this dual complementarity, we propose a hierarchical semantic-visual fusion strategy that reconstructs clear scenes via semantic guidance while preserving structural details, enabling more effective restoration for distant areas.

\noindent\textbf{Visible and near-infrared fusion.}
Although haze scattering primarily follows Mie theory with weak wavelength dependence \cite{narasimhan2002vision}, near-infrared light undergoes less attenuation than visible light due to its lower scattering and absorption coefficients. This makes it more effective in penetrating haze and preserving distant scene content. Hence, numerous methods exploit near-infrared images to assist haze removal \cite{feng2013near, jang2017colour, li2020spectrum, awad2019adaptive, jung2020unsupervised, yang2023detail, deng2020deep}, with many works formulating the task as a visible and near-infrared image fusion problem to enhance scene visibility \cite{lex2009color, ahn2011haze, ma2022swinfusion, zhao2022efficient, liu2022learning, zhu2023near, zhao2023cddfuse}. For example, Liu \emph{et al.} proposed jointly considering pixel alignment and fusion of visible and near-infrared images to solve the slight misalignment across modalities and enhance the details in long-range haze images \cite{liu2023lrinet}. While these methods have improved texture visibility, they often overlook the entangled haze present in visible images, leading to residual haze and limited contrast enhancement. Moreover, most approaches rely solely on low-level structural fusion, without exploring high-level semantic consistency across modalities. In this work, we go beyond low-level fusion by introducing a hierarchical semantic-visual fusion framework that jointly addresses haze elimination and structural recovery. Specifically, we leverage high-level semantic consistency across modalities as a powerful prior to guide clear scene reconstruction, thereby effectively eliminating residual haze and improving overall scene fidelity under severe long-range haze degradation.

\noindent\textbf{Semantic-guided restoration.}
The past few years have witnessed the development and effectiveness of semantic-guided restoration methods \cite{ren2018deep, zhang2021semantic, hong2022sg, liang2022semantically, wang2023uscformer, wu2024towards}. The key idea of these methods is to utilize semantic information as an auxiliary prior to facilitate adaptive image restoration at the feature or loss level. For example, Zhang et al. \cite{zhang2021semantic} generated segmentation maps through a multitask framework to facilitate the dehazing process through feature fusion. Liang et al. \cite{liang2022semantically} proposed a semantic brightness consistency loss to maintain scene details during low-light image enhancement. However, most existing methods rely on the semantics directly estimated from degraded inputs, which becomes particularly unreliable under severe degradation such as long-range haze. This severely limits their effectiveness in supporting subsequent restoration. In contrast, we propose to utilize near-infrared to assist semantic prior prediction. Motivated by the observation that semantics remains consistent across modalities, we align intrinsic semantic features between visible and near-infrared inputs to obtain robust semantic representations through the proposed intrinsic semantic alignment module, to facilitate the subsequent reconstruction process.

\section{Hierarchical Semantic-Visual Fusion}
\label{sec:method}

\begin{figure*}[t]
	\centering
	\includegraphics[width=\linewidth]{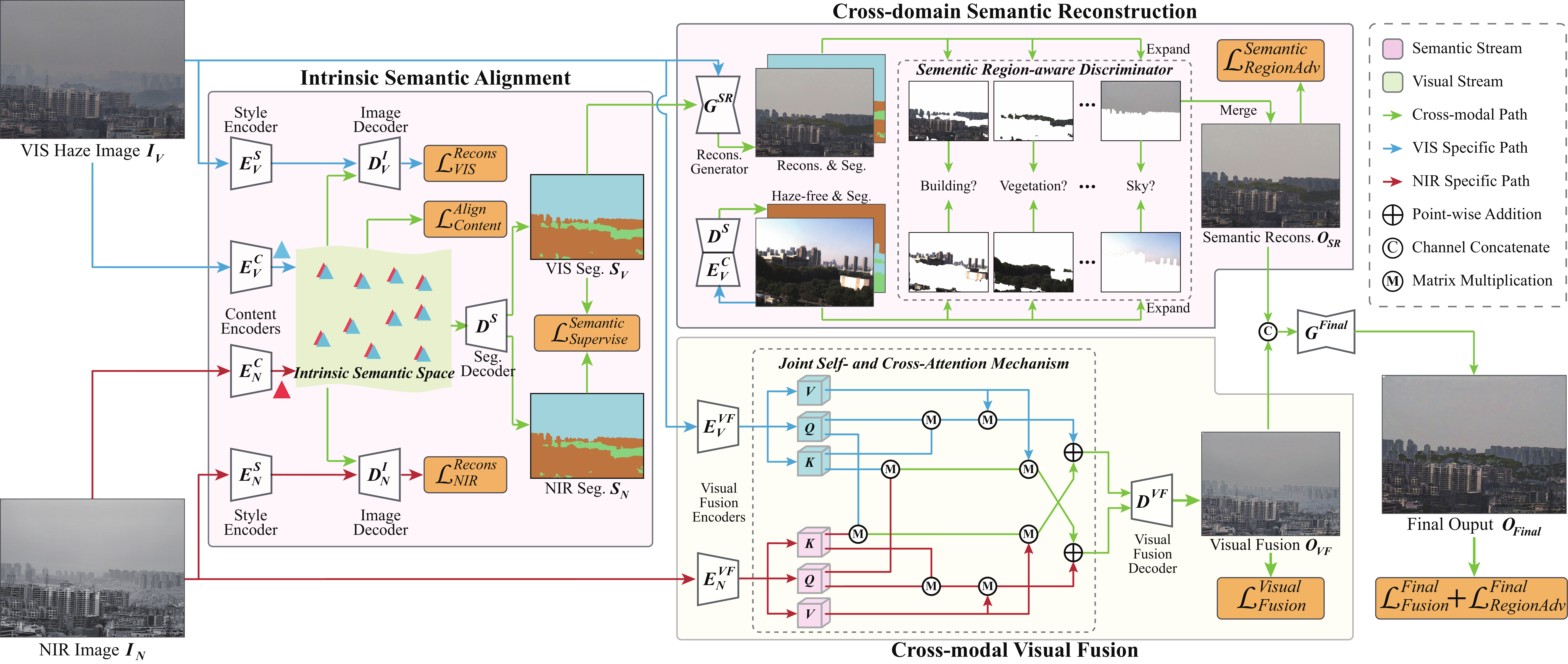}
	\caption{Overview of the proposed HSVF framework. The proposed HSVF framework consists of two streams: a semantic stream and a visual stream. The semantic stream aligns multimodal inputs in an intrinsic semantic space to extract precise, modality-invariant semantics in the intrinsic semantic alignment module. The extracted semantics then act as intermediate bridges to transfer clear weather domain knowledge and guide haze-free scene generation via a semantic region-aware discriminator in cross-domain semantic reconstruction module. The visual stream complements this by transferring fine structural details from the haze-resilient near-infrared modality utilizing joint self- and cross-attention within the cross-modal fusion module.
	}
	\label{fig_framework}
\end{figure*}

\begin{figure*}[t]
	\centering
	\includegraphics[width=\linewidth]{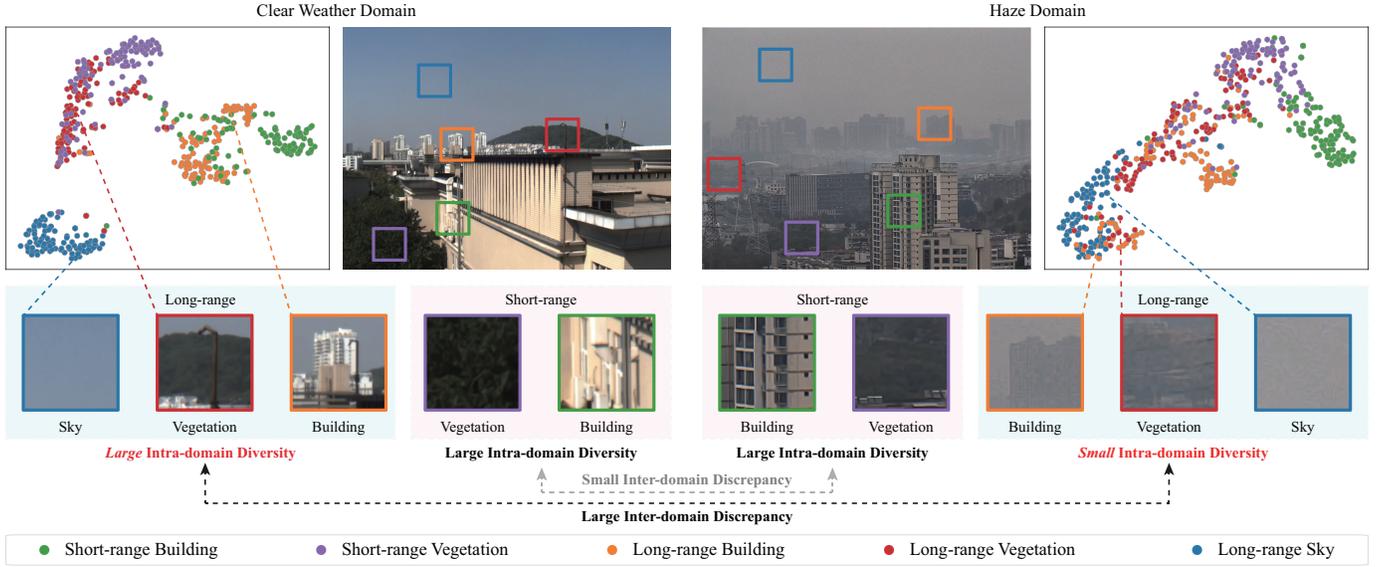}
	\caption{Illustration of the motivation for introducing semantics in long-range haze removal. We present t-SNE visualizations of semantic patches extracted from clear and hazy images. We first observe that the inter-domain gap between hazy and clear domains is especially pronounced in distant regions, where haze severely suppresses structural cues and semantic integrity, while short-range regions remain relatively consistent. Furthermore, in clear weather, semantic regions (e.g., sky, vegetation) form distinct and well-separated clusters in the t-SNE feature space, reflecting high intra-domain diversity, where each category retains its own structural and appearance patterns. In contrast, under haze conditions, features from different semantic categories become entangled and less distinguishable, especially in distant areas, indicating that semantic boundaries are severely blurred. These observations suggest that high-level semantics, as a discriminative prior, can effectively reintroduce structural separation across degraded regions, thereby guiding clearer scene reconstruction under severe haze.
	}
	\label{fig_motivation}
\end{figure*}

\noindent\textbf{Motivation.} 
In this work, we mainly focus on the challenging task of real-world long-range haze removal. The crux lies in two aspects: severe haze degradation and the loss of scene details, both of which would corrupt discriminative feature representations and significantly increase the difficulty of restoring distant scenes. To tackle this, we leverage the strong haze-penetration capability of near-infrared imagery to recover both low-level structural textures and high-level semantic cues. From a multimodal fusion perspective, these two types of information jointly serve as complementary priors to guide the restoration of clear and detailed long-range scenes. Hence, we introduce HSVF, a hierarchical semantic-visual fusion framework comprising two complementary streams: a semantic stream and a visual stream. An overview is shown in Fig. \ref{fig_framework}. Specifically, the semantic stream focuses on reconstructing clear scenes under modality-shared high-level semantic guidance, while the visual stream concentrates on retrieving and preserving low-level image structures and textures from both modalities, ensuring restoration fidelity.

\subsection{Semantic-guided Clear Scene Reconstruction}
\noindent\textbf{How semantic enables efficient scene reconstruction?} We first analyze how semantic information can support the reconstruction of long-range scenes. Figure \ref{fig_motivation} presents clear and hazy images alongside t-SNE visualizations of features from various semantic patches. We observe that the domain gap between clear and hazy images is especially pronounced in distant areas, where haze severely suppresses structural cues and semantic integrity. In contrast, near-range regions remain relatively consistent, and this discrepancy becomes more significant as the distance increases. This highlights the urgency to focus on reconstructing severely degraded distant regions. In addition, we find that semantic features in clear weather exhibit strong class separability. Regions such as sky, vegetation, and buildings form distinct and well-isolated clusters in the feature space, each maintaining unique structural and appearance patterns. However, under hazy conditions, these features become entangled and less distinguishable, particularly in distant regions where haze blurs structural cues. This blending of category features reduces representation discriminability and impairs reliable reconstruction. In light of this, high-level semantics can serve as a discriminative prior that reintroduces categorical separation and guide clearer, more structured scene reconstruction in severely degraded areas.

\noindent\textbf{Modality-shared intrinsic semantic alignment.}
As demonstrated in the preceding analysis, high-quality semantics could play a vital role in clear scene reconstruction, particularly in distant regions. However, long-range haze not only causes severe visual degradation, but also hampers discriminative feature extraction, thereby reducing semantic prediction accuracy \cite{sakaridis2018semantic, ma2022both} and ultimately compromising reconstruction quality in distant scenes. To obtain reliable semantics, we propose to jointly leverage both visible and haze-robust near-infrared spectra, rather than relying solely on degraded visible images.

Our key observation is that, despite large differences in visual appearance, visible and near-infrared images share consistent intrinsic scene semantics. For instance, a region labeled as “sky” retains its semantic identity across modalities, regardless of modality-induced appearance differences. The same holds for other categories such as buildings and vegetation. Based on this, we propose to disentangle features into three spaces: two modality-specific style spaces that capture appearance variations, and one modality-invariant semantic space that encodes intrinsic category information. By isolating semantics within this shared latent space, we suppress modality and haze interference, enabling more accurate semantic extraction.

Specifically, given a pair of visible and near-infrared images $I_{V}$ and $I_{N}$, we first employ two style encoders, $E^{S}_{V}$ and $E^{S}_{N}$, to map the inputs into two modality-specific style spaces. In parallel, two content encoders, $E^{C}_{V}$ and $E^{C}_{N}$, embed the images into a shared, modality-invariant intrinsic semantic feature space. To achieve such disentangled representations, we apply two strategies: (1) intrinsic semantic feature alignment, which encourages shared semantic representations across visible and near-infrared modalities, and (2) identity image reconstruction, which ensures content preservation. The feature alignment is enforced via a feature consistency loss:
\begin{equation}\
	\begin{aligned}
		\resizebox{!}{0.9\height}{$ \mathcal{L}^{Align}_{Content} = \Vert E^{C}_{V}(I_{V}) - E^{C}_{N}(I_{N}) \Vert_1. $}
		\label{eq2}
	\end{aligned}
\end{equation}
The aligned features provide a unified representation of modality-invariant scene semantics. Accordingly, we adopt a shared segmentation decoder $D^{S}$ supervised by a semantic segmentation loss defined as:
\begin{equation}\
	\begin{aligned}
		\resizebox{!}{0.9\height}{$ \mathcal{L}^{Seman}_{Super} = - {\sum_{n}^{}}  S_{GT}^{(n)}log(D^{S}(E^{C}_{V}(I_{V}))^{(n)}) $}\\
		\resizebox{!}{0.9\height}{$- \sum_{n}  S_{GT}^{(n)}log(D^{S}(E^{C}_{N}(I_{N}))^{(n)}), $}
		\label{eq3}
	\end{aligned}
\end{equation}
where $n$ denotes the semantic class index, and $S_{GT}^{(n)}$ is the corresponding ground truth for the $n$-th class.

As the disentangling process may lead to unintended information loss, we further introduce an image reconstruction loss to preserve comprehensive scene content. Specifically, we perform cross-modal image reconstruction by exchanging intrinsic semantic features between modalities and minimizing a corresponding reconstruction loss:
\begin{equation}\
	\begin{aligned}
		\resizebox{!}{0.9\height}{$ \mathcal{L}^{Recons}_{VIS} = \Vert D^{I}_{V}(E^{S}_{V}(I_{V}), E^{C}_{N}(I_{N}))  - I_{V} \Vert_1, $}
		\label{eq4}
	\end{aligned}
\end{equation}
\begin{equation}\
	\begin{aligned}
		\resizebox{!}{0.9\height}{$ \mathcal{L}^{Recons}_{NIR} = \Vert D^{I}_{N}(E^{S}_{N}(I_{N}), E^{C}_{V}(I_{V}))  - I_{N} \Vert_1, $}
		\label{eq5}
	\end{aligned}
\end{equation}
where $D^{I}_{V}$ and $D^{I}_{N}$ are image decoders for reconstructing visible and near-infrared (NIR) images from exchanged intrinsic semantics and style features. Hence, the overall loss of intrinsic semantic alignment module is formulated as:
\begin{equation}\
	\begin{aligned}
		\resizebox{!}{0.9\height}{$ \mathcal{L}^{Seman}_{Align} = \mathcal{L}^{Align}_{Content} + \mathcal{L}^{Seman}_{Super} + \mathcal{L}^{Recons}_{VIS} + \mathcal{L}^{Recons}_{NIR}. $}
		\label{eq6}
	\end{aligned}
\end{equation}
By jointly performing semantic feature alignment and identity-preserving image reconstruction, the proposed intrinsic semantic alignment module captures accurate semantics while suppressing haze interference, thereby providing reliable priors for subsequent scene reconstruction.

\noindent\textbf{Cross-domain semantic reconstruction.}
To address real-world long-range haze degradation, it is essential to transfer knowledge from haze-free domains to hazy counterparts. However, severe haze degradation often leads to substantial appearance discrepancies, rendering traditional direct image- or feature-level alignment and transfer strategies less effective. In this work, we introduce high-level semantics as an intermediate bridge between clear and hazy scenes to facilitate robust cross-domain knowledge transfer. Building on this, we propose a cross-domain semantic reconstruction module, in which a semantic region-aware discriminator guides the model to learn category-consistent and haze-free representations, thereby enabling accurate and faithful scene restoration under severe long-range haze degradation.

Specifically, we first generate the reconstruction results $O_{SR}$ via reconstruction generator $G_{SR}$. The corresponding clear image $I_{C}$ and its semantic segmentation $S_{C}$ are obtained from the haze-free domain via the previously introduced $E^{D}_{V}$ and $D^{S}$ in the intrinsic semantic alignment module. We then expand both $O_{SR}$ and $I_{C}$ into a set of category-aware regions (e.g., building, vegetation, sky), and apply a set of class-wise discriminators to encourage the generator to capture and translate category-specific discriminative features, thereby enhancing clear scene reconstruction. The semantic reconstruction loss is defined as:
\begin{equation}\
	\begin{aligned}
		\resizebox{!}{0.9\height}{$ \mathcal{L}^{Semantic}_{RegionAdv} = \mathbb{E}_{I_{C}}[{\textstyle \sum_{n}^{}} M_{n}^{I_C}log(D_{n}(I_{C}))] $}\\
		\resizebox{!}{0.9\height}{+ $\mathbb{E}_{O_{SR}}[{\textstyle \sum_{n}^{}} M_{n}^{  O_{SR}}log(1-D_{n}(O_{SR}))]. $}
		\label{eq7}
	\end{aligned}
\end{equation}
where $D_{n}$ denotes the discriminator corresponding to the $n$-th semantic category, and $M_{n}^{I_C}$, $M_{n}^{O_{SR}}$ are binary masks indicating the spatial regions of class $n$ in the clear image $I_C$ and the reconstruction output $O_{SR}$, respectively. Each $D_{n}$ focuses on learning category-specific discriminability to ensure semantic fidelity in the reconstructed scene. Compared to conventional direct image- or feature-level translation, our semantic region-aware discriminator explicitly incorporates semantics as an intermediate bridge by decomposing images into category-specific regions. Each semantic region is supervised by a dedicated discriminator, enabling fine-grained and class-consistent knowledge transfer across hazy and clear domains. This region-level adversarial learning not only mitigates domain discrepancies more effectively but also promotes the reconstruction of clear scenes under severe haze degradation.

\subsection{Complementary Visual Texture Fusion}
The fusion stream is designed to restore fine scene structures severely affected by long-range haze. Near-infrared images exhibit inherent resilience to haze and preserve richer structural details in distant regions. We leverage this property by transferring complementary visual features from the near-infrared modality to enhance the visible haze image. To effectively capture and integrate multimodal structural cues, we propose a cross-modal visual fusion module equipped with joint self- and cross-attention mechanisms. This enables robust modeling of modality-specific and complementary dependencies, facilitating the recovery of long-range textures and structures.

\noindent\textbf{Cross-modal visual fusion.}
The cross-modal visual fusion module is designed to preserve and enhance complementary structural information from both visible and near-infrared modalities. To achieve this, we disentangle features into modality-invariant and modality-specific components, and fuse them for richer cross-modal representation. Specifically, we first extract shallow features from both modalities and project them into query ($Q_{V}$, $Q_{N}$), key ($K_{V}$, $K_{N}$), and value ($V_{V}$, $V_{N}$) embeddings. The complementary representations $F_{V}$ and $F_{N}$ are then obtained by jointly attending to modality-specific and invariant features through self- and cross-attention:
\begin{equation}\
	\begin{aligned}
		\resizebox{!}{0.9\height}{$ F_{V} = \textrm{softmax}({{Q}_{V}{K}^T_{V}}/{\sqrt{d}}){V}_{V} +\textrm{softmax}({{Q}_{V}{K}^T_{N}}/{\sqrt{d}}){V}_{N}, $}
		\label{eq8}
	\end{aligned}
\end{equation}
\begin{equation}\
	\begin{aligned}
		\resizebox{!}{0.9\height}{$ F_{N} = \textrm{softmax}({{Q}_{N}{K}^T_{N}}/{\sqrt{d}}){V}_{N} +\textrm{softmax}({{Q}_{N}{K}^T_{V}}/{\sqrt{d}}){V}_{V}, $}
		\label{eq9}
	\end{aligned}
\end{equation}
where $d$ represents the feature dimension, ${K}^T_{V}$ and ${K}^T_{N}$ represent the transpose of ${K}_{V}$ and ${K}_{N}$. The former and latter terms in each equation represent the intra-modal self-attention and inter-modal cross-attention, respectively. Through our joint self- and cross-attention mechanism, modal-relevant and specific features are disentangled and fused adaptively for complementary feature representation. This design better preserves comprehensive visual structures from both modalities. We utilize image fusion loss \cite{ma2022swinfusion} for optimization:
\begin{equation}\
	\begin{aligned}
		\resizebox{!}{0.9\height}{$ \mathcal{L}^{Visual}_{Fusion} = \mathcal{L}^{Visual}_{ssim} + \mathcal{L}^{Visual}_{text} + \mathcal{L}^{Visual}_{int}, $}
		\label{eq10}
	\end{aligned}
\end{equation}
where the right three components correspond to SSIM loss, texture loss, and intensity loss, respectively:
\begin{equation}\
	\begin{aligned}
		\resizebox{!}{0.9\height}{$ \mathcal{L}^{Visual}_{ssim}=1-ssim(O_{VF},I_{V})+1-ssim(O_{VF},I_{N}), $}
		\label{eq11}
	\end{aligned}
\end{equation}
\begin{equation}\
	\begin{aligned}
		\resizebox{!}{0.9\height}{$ \mathcal{L}^{Visual}_{text} ={\textstyle \frac{1}{HW}}\left | \left | \nabla   O_{VF} - max(\left | \nabla I_{V} \right |,\left | \nabla   I_{N} \right | ) \right |  \right |_1, $}
		\label{eq12}
	\end{aligned}
\end{equation}
\begin{equation}\
	\begin{aligned}
		\resizebox{!}{0.9\height}{$ \mathcal{L}^{Visual}_{int} ={\textstyle \frac{1}{HW}}\left | \left |   O_{VF} - max(I_{V}, I_{N}) \right |  \right |_1, $}
		\label{eq13}
	\end{aligned}
\end{equation}
where $O_{VF}$ denotes the output of visual fusion module, $H$ and $W$ represent image height and width. Compared to existing methods, the proposed visual fusion module jointly embeds self- and cross-attention to disentangle and integrate complementary features, effectively transferring haze-resilient structures from near-infrared to visible images. This facilitates the recovery of fine details lost due to long-range haze.

\begin{figure*}[t]
	\centering
	\includegraphics[width=\linewidth]{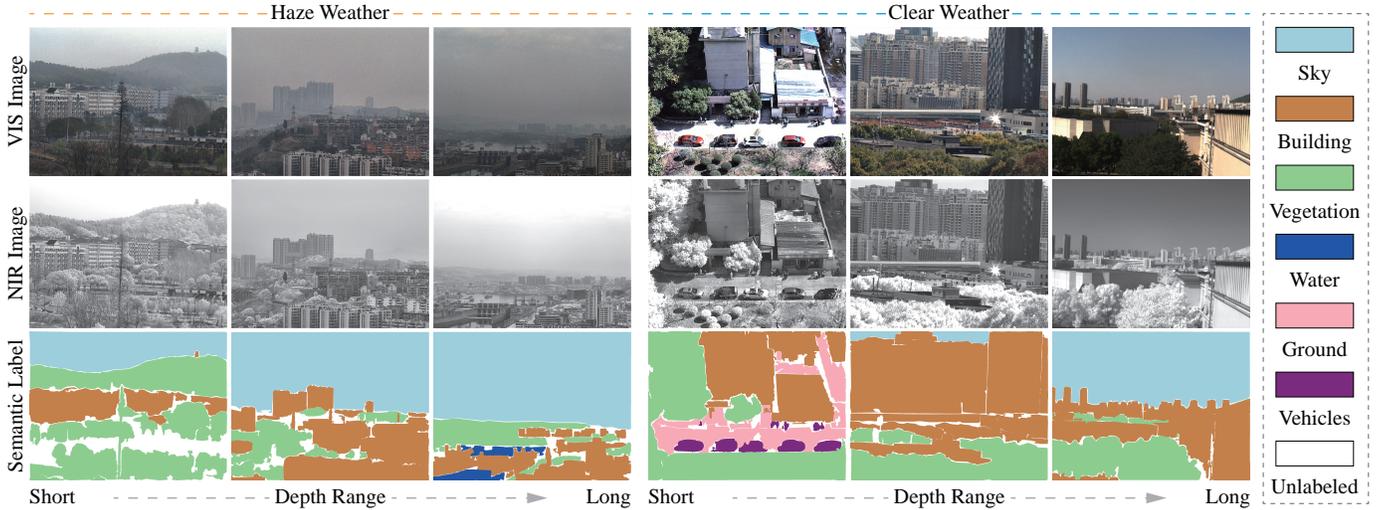}
	\caption{Examples of the proposed real-world pixel aligned multimodal haze dataset VNHD. From top to bottom are paired visible, near-infrared, and the corresponding semantic labels respectively. Our VNHD possesses samples with various imaging distances under both haze and clear conditions.}
	\label{fig_dataset}
\end{figure*}

\begin{table*}[t] \centering
	\caption{Quantitative comparison with existing state-of-the-art methods on VNHD, RGB-NIRScene \cite{brown2011multi}, and RANUS \cite{choe2018ranus} dataset. We compare the proposed HSVF with supervised dehazing method Dehamer \cite{guo2022image}, C2PNet \cite{zheng2023curricular}, DA-Clip \cite{luo2024controlling}, DCMPNet \cite{zhang2024depth}, AdaIR \cite{cui2025adair}; semi- or unsupervised dehazing methods PSD \cite{chen2021psd}, RIDCP \cite{wu2023ridcp}, UME-Net \cite{sun2024unsupervised}, ODCR \cite{wang2024odcr}, D4+ \cite{yang2024robust}; and fusion-based methods SwinFusion \cite{ma2022swinfusion}, MHLP \cite{yang2023detail}, CDDFuse \cite{zhao2023cddfuse}, EMMA \cite{zhao2024equivariant}. The best and second-best scores are highlighted in \textcolor{red}{red} and \textcolor{orange}{orange}.}
	\label{tab1}
	\setlength{\tabcolsep}{4mm}
	\scalebox{1}
	{\begin{tabular}{c|l|l|cc|cc|cc}
			\toprule[1.5pt]
			\multirow{2}{*}{\shortstack{Type}} & \multirow{2}{*}{\shortstack{Methods}}  &  \multirow{2}{*}{\shortstack{Publication}}     & 
			\multicolumn{2}{c|}{VNHD Dataset}  & \multicolumn{2}{c|}{NIRScene Dataset} & 
			\multicolumn{2}{c}{RANUS Dataset} \\
			&&&FADE $\downarrow$ & NIQE $\downarrow$& FADE $\downarrow$ & NIQE $\downarrow$& FADE $\downarrow$ & NIQE $\downarrow$ \\
			\midrule[1.5pt]
			--- & Haze                             & ---       & 1.5043        & 5.3483        & 1.3677        & 3.1259 & 1.4848 & 4.5215 \\\midrule[0.5pt]
			\multirow{4}{*}{\shortstack{Supervised}}
			&Dehamer        & 2022 CVPR & 1.0280        & 5.2869        & 0.9469        & 3.5914 & 0.7558 & 3.9138 \\
			&C2PNet  & 2023 CVPR & 1.1438        & 5.1797        & 0.9997        & 3.0717 & 0.8082 & 4.2728 \\
			&DA-Clip    & 2024 ICLR & 1.1157        & 4.3243        & 0.8586        & 3.0710 & 1.0583 & 3.8716 \\
			&DCMPNet       & 2024 CVPR & 0.9145        & 4.7998        & 0.8602        & 3.6642 & \textcolor{orange}{0.5899} & 4.0899 \\
			&AdaIR                      & 2025 ICLR & 0.9276        & 4.5132        & 0.8515        & 3.0613 & 0.9598 & 4.3913 \\\midrule[0.5pt]
			\multirow{4}{*}{\shortstack{Semi- or\\Un-supervised}}
			&PSD        & 2021 CVPR & 0.7696        & 4.7834        & 0.6968        & 4.4377 & 0.6709 & 5.7052 \\
			&RIDCP      & 2023 CVPR & 0.9190        & 6.2556        & 0.6740        & \textcolor{orange}{2.9372} & 0.7168 & 3.7162 \\
			&UME-Net                    & 2024 PR   & 1.1228        & \textcolor{orange}{4.4014}      & 0.6218 & 3.8396 & 0.6960 & 3.7125 \\
			&ODCR       & 2024 CVPR & 0.8358        & 4.9224        & 0.6631        & 3.0488 & 0.8581 & 3.8910 \\
			&D4+                       & 2024 IJCV &\textcolor{red}{0.6594}&4.5419&\textcolor{red}{0.3379}& 3.3903 & \textcolor{red}{0.5662} & 4.5241 \\\midrule[0.5pt]
			\multirow{4}{*}{\shortstack{Fusion-based}}
			&SwinFusion & 2022 JAS  & 1.1087        & 4.4248        & 1.1726        & 3.1645 & 1.1178 & 3.5072 \\
			&MHLP        & 2023 IF   & 0.9007        & 5.5672        & 0.7580        & 3.3802 & 0.7717 & \textcolor{orange}{3.2535} \\
			&CDDFuse     & 2023 CVPR & 1.3800        & 4.8123        & 1.1435        & 3.2991 & 0.9780 & 4.5451 \\
			&EMMA        & 2024 CVPR & 1.0742        & 4.6887        & 1.0009        & 4.3317 & 0.9213 & 4.7123 \\\midrule[0.5pt]
			--- &HSVF                              & ---       &\textcolor{orange}{0.7392}&\textcolor{red}{4.2162}&\textcolor{orange}{0.5955}&\textcolor{red}{2.7038}&0.6365&\textcolor{red}{3.2401} \\
			\bottomrule[1.5pt]
	\end{tabular}} 
\end{table*}

\noindent\textbf{Final output generation and overall loss.} 
Finally, the output image $O_{Final}$ is obtained by fusing $O_{SR}$ and $O_{VF}$ via final generator $G^{Final}$. To ensure that $O_{Final}$ inherits both semantic clarity and structural richness, we apply a region-adversarial loss $\mathcal{L}^{Final}_{RegionAdv}$ and a fusion consistency loss $\mathcal{L}^{Final}_{Fusion}$, which follow the same formulation as $\mathcal{L}^{Semantic}_{RegionAdv}$ and $\mathcal{L}^{Visual}_{Fusion}$. The overall loss is defined as follows:
\begin{equation}\
	\begin{aligned}
		\resizebox{!}{0.9\height}{$ \mathcal{L}_{Total} = \lambda\mathcal{L}^{Semantic}_{Align} + \alpha\mathcal{L}^{Semantic}_{RegionAdv} + \beta\mathcal{L}^{Visual}_{Fusion}$}\\
		\resizebox{!}{0.9\height}{$ + \alpha_1\mathcal{L}^{Final}_{RegionAdv} + \beta_1\mathcal{L}^{Final}_{Fusion}. $}
		\label{eq14}
	\end{aligned}
\end{equation}
Through Eq.\ref{eq14}, the proposed HSVF effectively ensures both haze-free scene reconstruction and the recovery of lost structural details, enabling robust long-range haze removal.

\subsection{VNHD Dataset}
\label{sec:dataset}
\noindent\textbf{Motivation.} 
Although numerous datasets have been proposed for image dehazing \cite{li2018benchmarking, ancuti2019dense, liu2021synthetic} and visible-near-infrared fusion \cite{brown2011multi, luthen2017rgb, choe2018ranus}, there remains a lack of datasets tailored to real-world long-range haze removal, particularly in multimodal settings. Existing benchmarks for dehazing often rely on synthetic haze and focus on short-range scenarios, while commonly used RGB-NIR fusion datasets are not specifically designed for haze conditions. Some studies have attempted to use RGB-NIRScene \cite{brown2011multi} for haze removal, but this dataset was acquired through sequential imaging, leading to slight pixel misalignment in dynamic scenes and limiting its effectiveness for image fusion. To address these limitations, we introduce VNHD, a novel real-world, pixel-aligned visible-near-infrared haze dataset specifically constructed for long-range imaging. VNHD is designed to support multimodal scene understanding and provide a reliable benchmark for future research in image restoration and parsing under long-range haze scenarios.

\noindent\textbf{Dataset acquisition and characteristics.} 
We construct VNHD using the JAI FS-3200D-10GE dual-band camera, which features a built-in coaxial beam-splitting system for hardware-level pixel alignment between visible and near-infrared (NIR) channels. Each image pair in the VNHD dataset consists of a 3-channel RGB visible image and a single-channel NIR image, captured at a native resolution of 2048×1536 pixels and stored in 8-bit PNG format. No post-alignment, resizing, or cropping is applied to preserve raw imaging fidelity. The VNHD dataset covers hazy and clear weather conditions, and comprises 2961 real-world image pairs: 1519 for haze and 1442 for clear weather (examples shown in Fig. \ref{fig_dataset}). Data collection was conducted across different times of day and locations to ensure scene diversity and capture various real-world atmospheric conditions. VNHD includes scenes with depths ranging from 1 km to over 10 km, covering 500 images under 3 km (16.9\%), 1211 between 3-10 km (40.9\%), and 1250 over 10 km (42.2\%). This variation supports a comprehensive evaluation of visibility degradation under different depths. To mitigate sensor noise, especially in hazy conditions where illumination is low, we apply BM3D denoising \cite{dabov2007image}, ensuring high-quality images for evaluation.

\begin{figure*}[t]
	\centering
	\includegraphics[width=\linewidth]{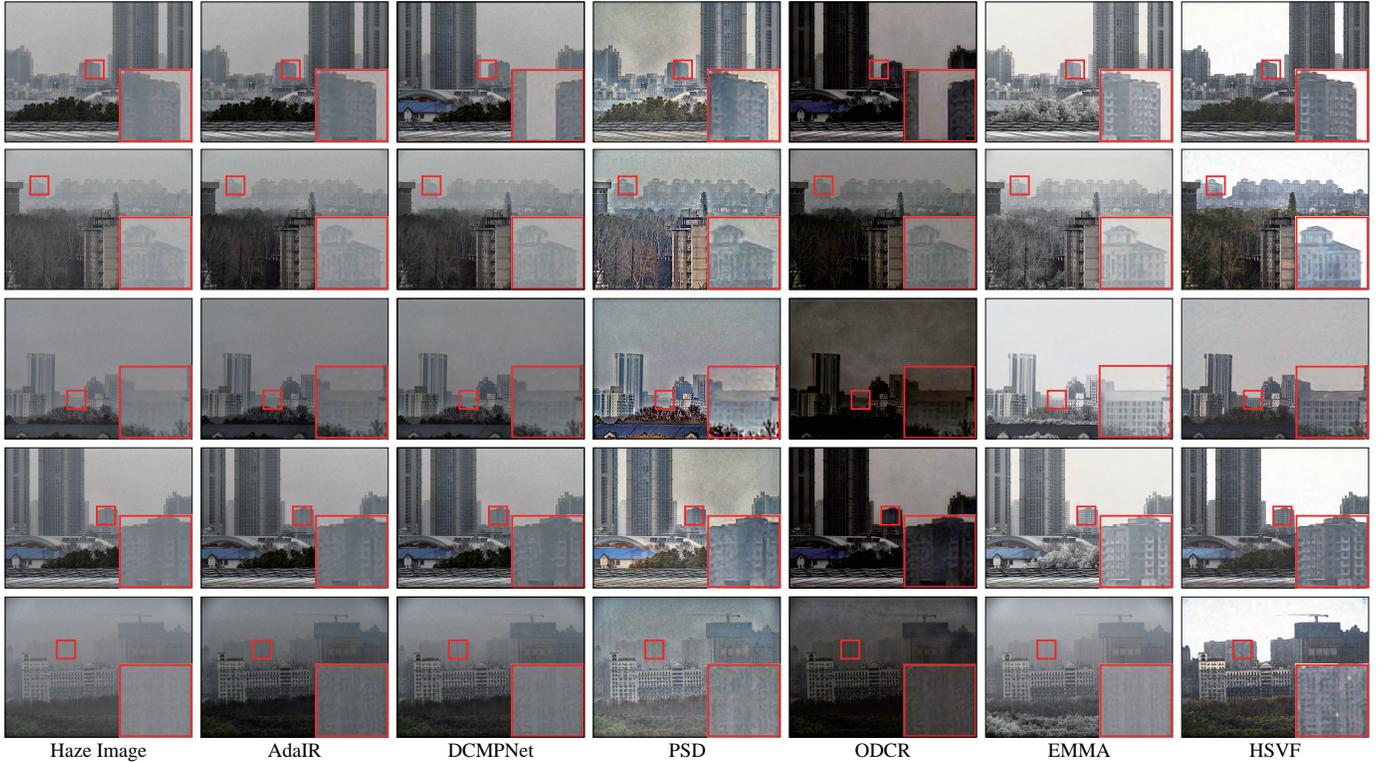}
	\caption{Comparison of real-world haze removal on the proposed VNHD dataset. We compare the proposed HSVF with supervised dehazing method AdaIR \cite{cui2025adair}, DCMPNet \cite{zhang2024depth}; semi- or unsupervised dehazing methods PSD \cite{chen2021psd}, ODCR \cite{wang2024odcr}; and fusion-based methods EMMA \cite{zhao2024equivariant}. }
	\label{fig_compare_VNHD}
\end{figure*}

\begin{table}[t] \centering
	\caption{Ablation study to the influence of each component in intrinsic semantic alignment module for semantic segmentation performance.}
	\label{tab_intrinsic_semantic}
	{\begin{tabular}{cccc|cc}
			\toprule[1.5pt]
			VIS        & NIR        & $\mathcal{L}^{Align}_{Content}$ & $\mathcal{L}^{Recons}$ & mIoU            & PixelACC          \\\midrule[1.5pt]
			\checkmark &            &                                 &                        & 0.7904          & 0.9862            \\
			& \checkmark &                                 &                        & 0.7949          & 0.9877            \\
			\checkmark & \checkmark &                                 &                        & 0.7973          & 0.9856            \\
			\checkmark & \checkmark & \checkmark                      &                        & 0.7989          & \textbf{0.9883}            \\
			\checkmark & \checkmark & \checkmark                      & \checkmark             & \textbf{0.8019} & 0.9882   \\
			\bottomrule[1.5pt]
	\end{tabular}}
\end{table}

\noindent\textbf{Segmentation annotation.} 
Moreover, we provide segmentation annotations for all visible-NIR image pairs. Given the focus on long-range scenarios, we select six representative categories frequently occurring in distant scenes: sky, ground, buildings, vegetation, water, and vehicles. Examples are shown in the bottom of Fig. \ref{fig_dataset}. We annotate 600 images for validation: 300 under haze and 300 under clear weather to ensure fair evaluation across atmospheric conditions. The remaining images are reserved for training, providing a solid foundation for developing segmentation models tailored to long-range scenes.

\begin{figure*}[t]
	\centering
	\includegraphics[width=\linewidth]{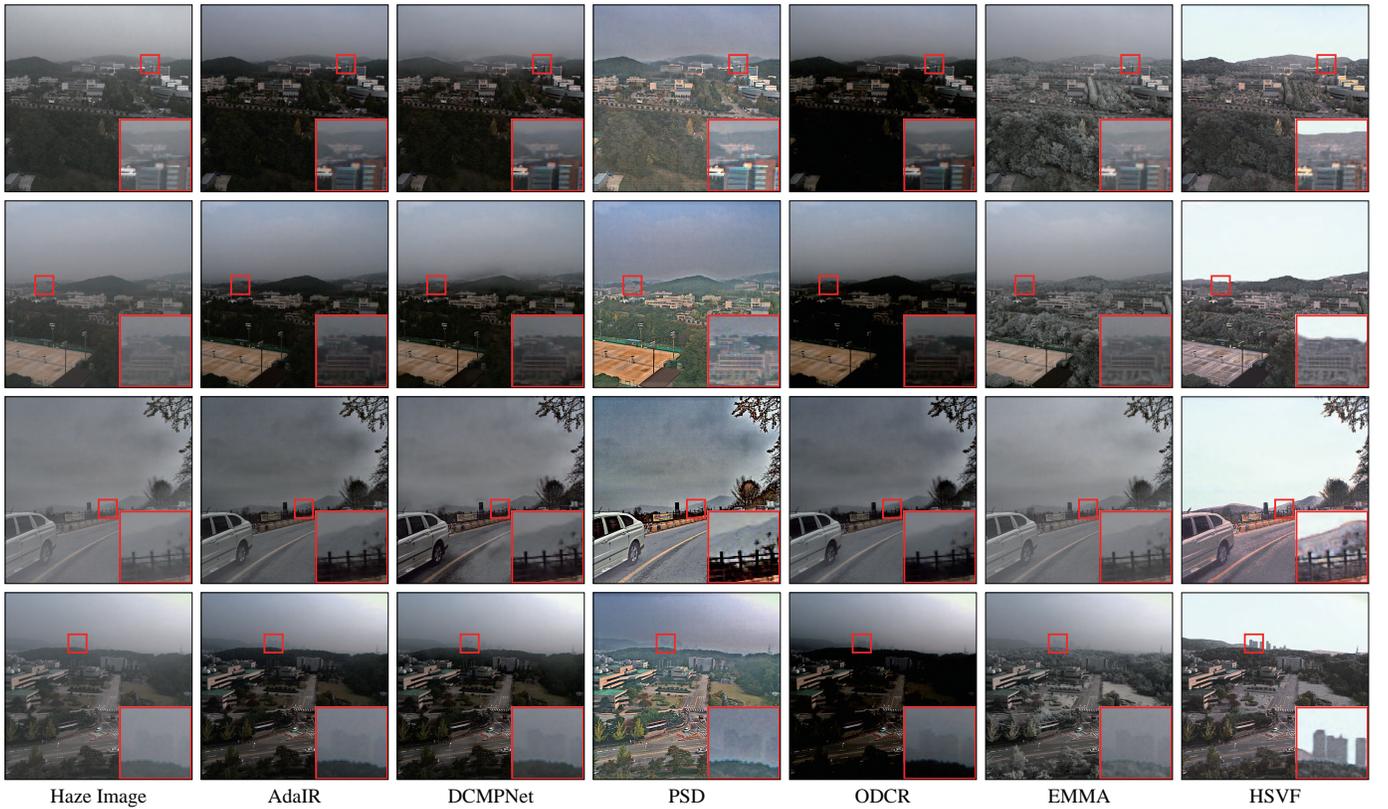}
	\caption{Comparison of real-world haze removal on RANUS dataset \cite{choe2018ranus}. We compare the proposed HSVF with supervised dehazing method AdaIR \cite{cui2025adair}, DCMPNet \cite{zhang2024depth}; semi- or unsupervised dehazing methods PSD \cite{chen2021psd}, ODCR \cite{wang2024odcr}; and fusion-based methods EMMA \cite{zhao2024equivariant}. }
	\label{fig_compare_RANUS}
\end{figure*}

\begin{figure*}[t]
	\centering
	\includegraphics[width=\linewidth]{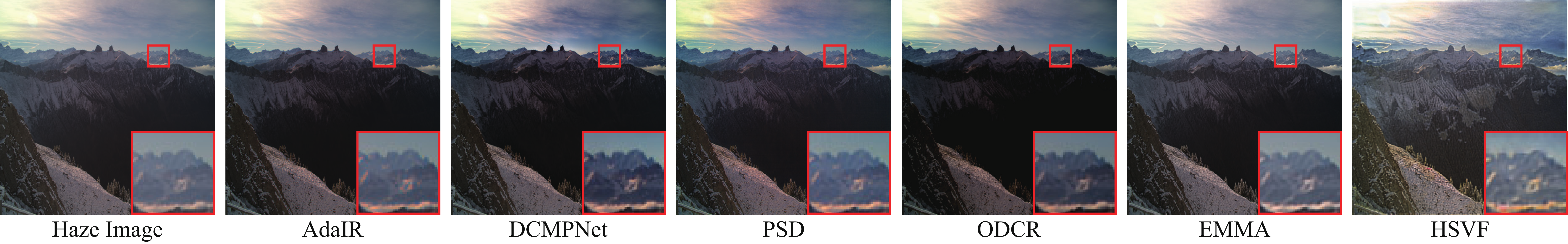}
	\caption{Comparison of real-world haze removal on RGB-NIRScene \cite{brown2011multi}. We compare the proposed HSVF with supervised dehazing method AdaIR \cite{cui2025adair}, DCMPNet \cite{zhang2024depth}; semi- or unsupervised dehazing methods PSD \cite{chen2021psd}, ODCR \cite{wang2024odcr}; and fusion-based methods EMMA \cite{zhao2024equivariant}. }
	\label{fig_compare_NIRScene}
\end{figure*}

\section{Experiments}
\label{sec:experiments}

\begin{table}[tb] \centering
	\caption{Analysis of different attention mechanisms for preservation of visual structures in visual appearance fusion. The evaluation metrics MI, VIF represent mutual information, visual information fidelity from the image fusion task, respectively.}
	\label{tab3}
	{\begin{tabular}{cc|ccc}
			\toprule[1.5pt]
			Self-Attention & Cross-Attention &  MI     & VIF    & $Q_{AB/F}$   \\\midrule[1.5pt]
			&             &  3.1493 & 0.8864 & 0.6131   \\
			\checkmark &             &  3.2535 & 0.9174 & 0.6263   \\
			& \checkmark  &  3.2534 & 0.9349 & 0.6332   \\
			\checkmark & \checkmark  &  \textbf{3.2548} & \textbf{0.9523} & \textbf{0.6419}   \\
			\bottomrule[1.5pt]
	\end{tabular}}
\end{table}

\subsection{Implementations and Experiments Setting}
\noindent\textbf{Implementation details.}
Our HSVF is implemented in PyTorch and trained on four RTX 3090 GPUs. We use HRNet \cite{wang2020deep} as the content encoders $E^C_V$ and $E^C_N$, and a modified U-Net \cite{ronneberger2015u} as the style encoders $E^S_V$ and $E^S_N$ in the intrinsic semantic alignment module. The semantic reconstruction generator adopts SPADE \cite{park2019semantic}. As for balanced weights, we empirically set $\lambda$, $\alpha$, $\beta$, $\alpha_1$, and $\beta_1$ as 1, 0.1, 0.01, 1, and 0.1. To ensure stable optimization, we pretrain each major module individually before joint fine-tuning. Specifically, the semantic alignment module is pretrained for 200 epochs using paired visible-NIR images and segmentation labels. The cross-domain semantic reconstruction and visual appearance fusion modules are both pretrained for 400 epochs using haze/haze-free images and modality pairs, respectively. Finally, the full framework is jointly finetuned for 200 epochs.

\noindent\textbf{Experiments settings.}
We compare the proposed HSVF with supervised dehazing method Dehamer \cite{guo2022image}, C2PNet \cite{zheng2023curricular}, DA-Clip \cite{luo2024controlling}, AdaIR \cite{cui2025adair}, DCMPNet \cite{zhang2024depth}; semi- or unsupervised dehazing methods PSD \cite{chen2021psd}, RIDCP \cite{wu2023ridcp}, UME-Net \cite{sun2024unsupervised}, D4+ \cite{yang2024robust}, ODCR \cite{wang2024odcr}; and fusion-based methods SwinFusion \cite{ma2022swinfusion}, MHLP \cite{yang2023detail}, CDDFuse \cite{zhao2023cddfuse}, EMMA \cite{zhao2024equivariant}. For fair comparisons, we have finetuned all methods (if possible) with codes provided by authors. We compare long-range haze removal performance between HSVF and other state-of-the-art methods on VNHD, RGB-NIRScene \cite{brown2011multi}, and RANUS \cite{choe2018ranus}.

\subsection{Comparison with State-of-the-art}
\noindent\textbf{Quantitative comparisons.}
We mainly utilize several non-reference metrics for quantitative comparison on real-world data. We adopt the Fog Aware Density Evaluator (FADE) \cite{choi2015referenceless} and NIQE \cite{mittal2012making} to evaluate long-range haze removal performance quantitatively, as shown in Tab. \ref{tab1}. From the results, we have three key observations. (1) Dehazing methods tend to achieve better FADE scores. This is because dehazing methods primarily focus on suppressing haze degradation and enhancing image contrast. However, due to the strong scattering effects in long-range haze conditions, most long-range scene detail is irreversibly lost. As a result, the restored images by dehazing methods tend to suffer from low sharpness and poor texture richness, leading to relatively higher NIQE scores and poorer perceived visual quality. (2) Fusion-based methods exhibit an opposite trend and tend to acquire better NIQE scores. By leveraging the near-infrared modality, which retains richer texture and structural information, fusion methods often produce images with sharper edges and richer textures, achieving better NIQE scores. Nevertheless, most fusion-based approaches primarily focus on integrating low-level structures without explicitly addressing the haze degradation embedded in visible images. As a result, residual haze often remains in the fusion outputs, resulting in relatively higher FADE scores and limited contrast improvement. (3) Among dehazing methods, semi-supervised and unsupervised approaches tend to outperform supervised methods in FADE. This phenomenon is mainly due to the significant domain gap between synthetic haze used for training and real-world haze scenarios. Supervised methods often struggle to generalize to real-world long-range haze conditions, leading to reduced dehazing effectiveness and higher FADE scores.

\begin{figure*}[t]
	\centering
	\includegraphics[width=\linewidth]{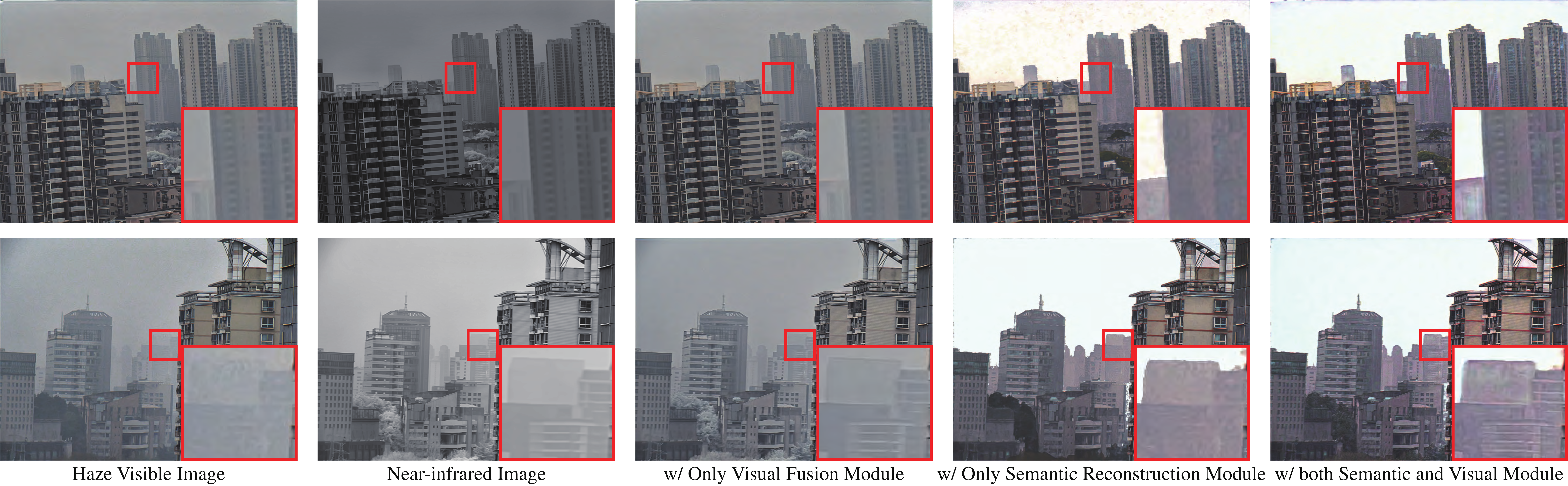}
	\caption{Ablation study of semantic reconstruction module and visual fusion module in HSVF framework. HSVF is capable of obtaining results with rich colors and high contrast from semantic reconstruction module, while retrieving the lost scene details caused by long-range haze via visual fusion module.}
	\label{fig_ablation_semantic_visual}
\end{figure*}

\begin{figure}[t]
	\centering
	\includegraphics[width=\linewidth]{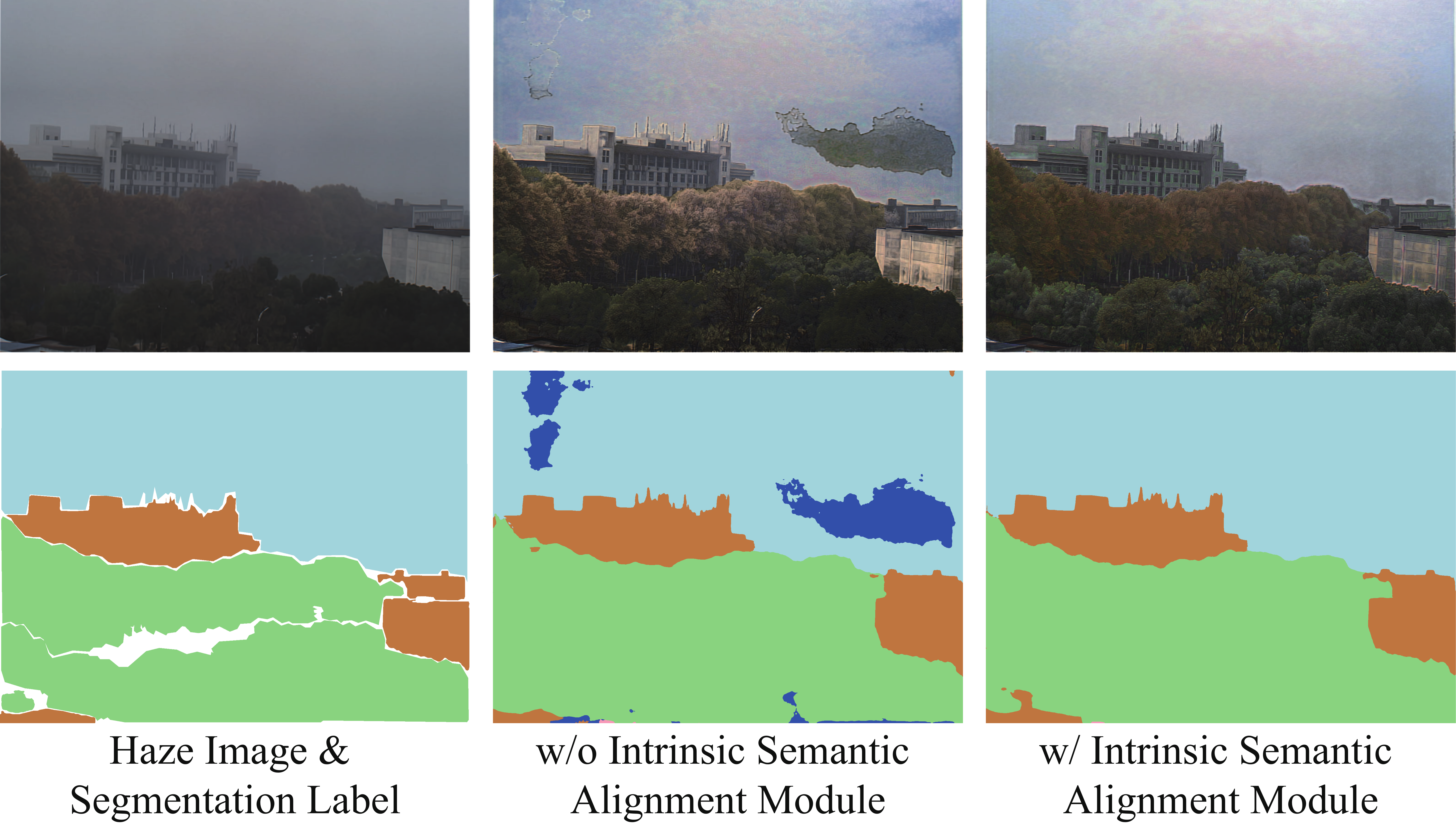}
	\caption{Ablation study of intrinsic semantic alignment module to long-range haze removal. Bottom to the top are segmentation and dehazing results.}
	\label{fig_ablation_intinsic_semantic}
\end{figure}

These observations collectively highlight two key conclusions: First, it is crucial to simultaneously address haze removal and multimodal detail enhancement when restoring long-range degraded scenes. Second, leveraging unsupervised learning paradigms to transfer knowledge from clear domains to real-world hazy conditions proves highly effective. Benefiting from these insights, our HSVF method jointly addresses clear scene reconstruction and detail retrieval, while further utilizing high-level semantic information as a bridge to facilitate robust knowledge transfer. Although DCMPNet \cite{zhang2024depth} achieves the second-best FADE score on RANUS dataset \cite{choe2018ranus}, our HSVF framework still consistently attains either the best or second-best performance across all metrics on other datasets. 

\begin{table}[t] \centering
	\caption{Quantitative comparison of results under different hyper-parameter setting in the final loss function.}
	\label{tab_parameter_analysis}
	\setlength{\tabcolsep}{4mm}
	\scalebox{1}
	{\begin{tabular}{ll|cc}
			\toprule[1.5pt]
			\multicolumn{2}{c|}{Parameter}           & FADE $\downarrow$ & NIQE $\downarrow$   \\\midrule[1pt]
			$\alpha_1$ = 0    &$\beta_1$ = 0.1           & 1.3208            & 3.9871              \\
			$\alpha_1$ = 0.01 &$\beta_1$ = 0.1           & 0.9088            & 4.0524              \\
			$\alpha_1$ = 0.1  &$\beta_1$ = 0.1           & 0.7456            & 4.3308              \\
			$\alpha_1$ = 1    &$\beta_1$ = 0.1           & 0.7392            & 4.2162              \\
			$\alpha_1$ = 10   &$\beta_1$ = 0.1           & 0.8410            & 4.5143              \\
			$\alpha_1$ = 10   &$\beta_1$ = 0             & 1.1323            & 4.5567              \\
			\bottomrule[1.5pt]
	\end{tabular}} 
\end{table}

\noindent\textbf{Qualitative comparisons.}
As for qualitative comparison in Fig. \ref{fig_compare_VNHD}, supervised dehazing method AdaIR \cite{cui2025adair} and DCMPNet \cite{zhang2024depth} fails to solve real-world long-range haze due to the huge gap between synthetic and real data. Semi-supervised PSD \cite{chen2021psd} can acquire dehazing images with higher contrast, however, the results may still suffer from severe loss of scene detail caused by long-range haze. Moreover, when encountering low-illumination entangled in long-range haze image, ODCR \cite{wang2024odcr} tends to severely darken the whole image. With haze-robust near-infrared preserving image structure of distant areas, fusion-based EMMA \cite{zhao2024equivariant} could acquire results with rich visual structures. However, solely focusing on fusion of low-level visual texture while neglecting the color and contrast restoration, the results of these methods may exhibit unexpected haze remaining and unrealistic color. Compared with these methods, results of the proposed HSVF possess not only higher contrast and satisfying scene color but also comprehensive visual structure thanks to hierarchical semantic-visual transference strategy. Furthermore, we also provide results on RGB-NIRScene and RANUS. As shown in Fig. \ref{fig_compare_NIRScene} and \ref{fig_compare_RANUS}, we could observe that the proposed HSVF could also achieve haze-free and rich texture results compared with other dehazing \cite{cui2025adair, wang2024odcr} and image fusion methods \cite{zhao2024equivariant}, which further demonstrate the effectiveness of the proposed HSVF framework for long-range haze removal.

\subsection{Ablation Study}

\noindent\textbf{Influence of semantic reconstruction and visual fusion.} As shown in Fig. \ref{fig_ablation_semantic_visual}, when the semantic reconstruction module is removed, the visual fusion module functions alone. In this case, the results mainly reflect the faithful fusion of textures and intensity information from the visible and near-infrared inputs. Benefiting from the haze-penetrating capability of near-infrared imaging, some distant structures are preserved and reflected in the fused outputs. However, due to the lack of clear scene reconstruction capability provided by the semantic reconstruction module, severe haze residuals persist, resulting in outputs with low contrast. Conversely, when the visual fusion module is removed and only the semantic reconstruction module remains active, the model primarily focuses on reconstructing a clear haze-free scene. While the global contrast is improved, the absence of fine-grained texture fusion leads to results with relatively limited texture richness, especially in regions where the visible input suffers from severe information loss. When both the semantic reconstruction and visual fusion module work together, the framework is able to simultaneously reconstruct a high-contrast clear scene and recover fine texture structures by integrating multimodal information. This joint effect leads to outputs with both enhanced contrast and enriched texture details. These results clearly validate the effectiveness of the proposed hierarchical semantic-visual fusion framework, which is designed to fully exploit strengths of high-level semantic reconstruction and low-level structural fusion.

\begin{figure*}[t]
	\centering
	\includegraphics[width=0.8\linewidth]{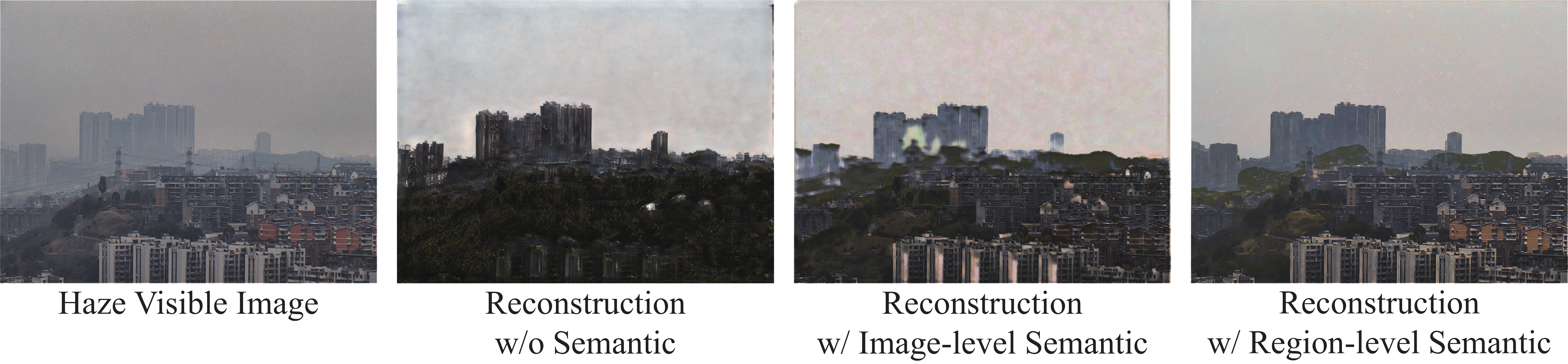}
	\caption{Analysis of different level of semantic guidance for scene reconstruction. Directly reconstructing clear scenes without semantic guidance often leads to content distortion, such as reconstructing buildings with vegetation textures. Incorporating image-level semantics enhances content invariance but still may introduce artifacts. The proposed region-level semantic guidance in semantic region-aware discriminator further improves realism and semantic consistency.}
	\label{fig_analysis_keycomponents}
\end{figure*}

\begin{figure*}[t]
	\centering
	\includegraphics[width=\linewidth]{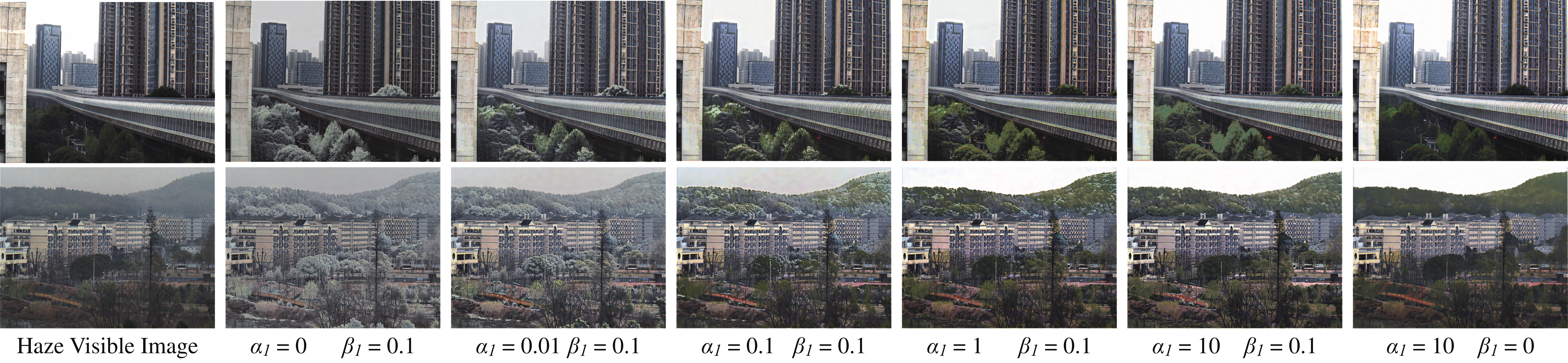}
	\caption{Analysis of how semantic and visual loss weight ratio influences the final long-range haze removing results. As the ratio increases, more realistic color and haze-free contrast are restored with less texture retrieval, and vice versa.}
	\label{fig_analysis_parameter}
\end{figure*}

\noindent\textbf{Influence of intrinsic semantic alignment module.}
The proposed intrinsic semantic alignment module aims at acquiring haze-robust semantics with near-infrared assistance; we first evaluate the influence of each component on segmentation performance within the intrinsic semantic alignment module, as shown in Tab. \ref{tab_intrinsic_semantic}. We can observe that utilizing both visible and near-infrared images is beneficial for facilitating segmentation compared with single-modal segmentation. And both $\mathcal{L}^{Recons}$ and $\mathcal{L}^{Align}_{Content}$ serve important roles in quantitative performance. The results in the third and fourth rows demonstrate the benefits brought by intrinsic feature alignment and image reconstruction respectively. By disentangling and aligning the intrinsic semantic features of visible and near-infrared images, the proposed intrinsic semantic alignment module is capable of extracting modality-shared intrinsic semantic feature representation that is invariant to modality and degradation, so as to acquire better segmentation results.

\begin{figure*}[t]
	\centering
	\includegraphics[width=\linewidth]{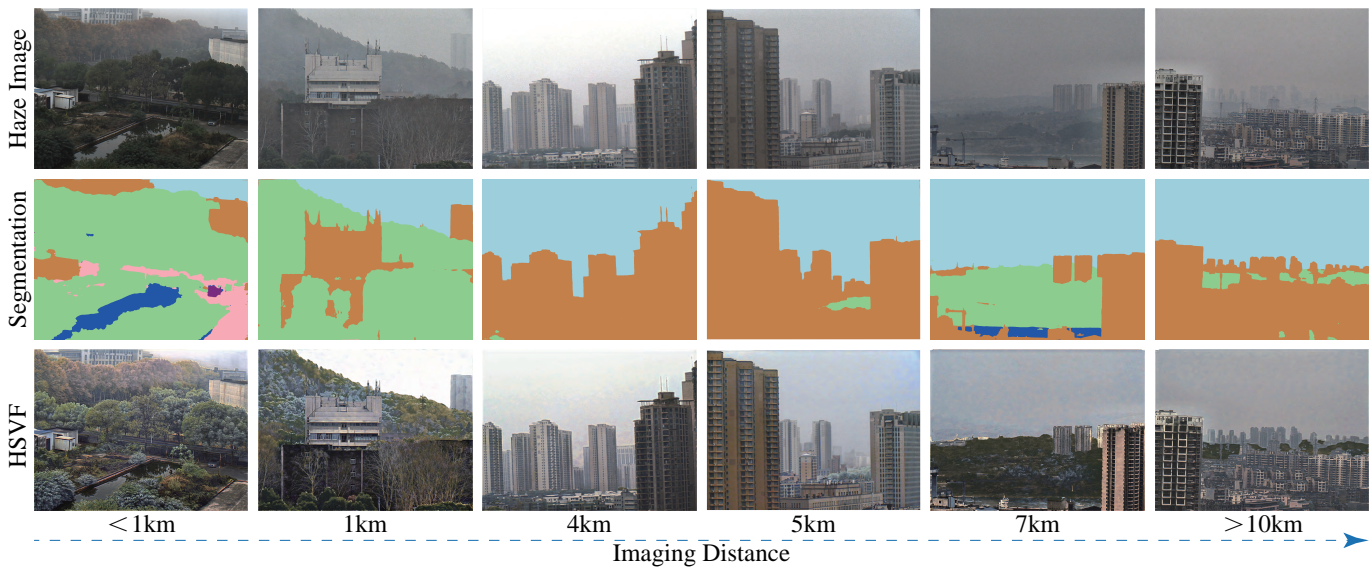}
	\caption{Visualization of haze removal result of HSVF under different depth ranges. From left to right, the depth range increases from lower than 1km to over 10 km. We can observe that the proposed HSVF can effectively handle haze removal tasks under scenes across various depth ranges with satisfying results.}
	\label{fig_analysis_imagingdistance}
\end{figure*}

\begin{figure}[t]
	\centering
	\includegraphics[width=\linewidth]{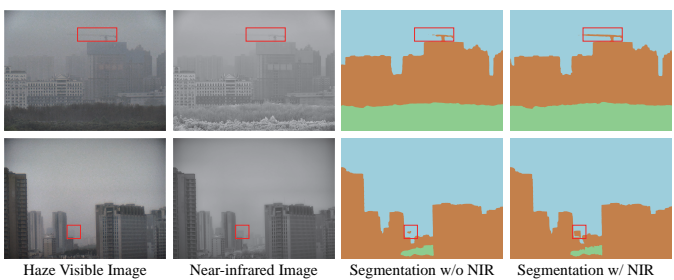}
	\caption{Analysis of how the near-infrared spectrum facilitate long-range scene content segmentation.}
	\label{fig_analysis_modalitysemantic}
\end{figure}

Moreover, we provide visualized results to analyze the effect of the intrinsic semantic alignment module on the final long-range haze removal performance. As shown in Fig. \ref{fig_ablation_intinsic_semantic}, our observations reveal that, in the absence of an intrinsic semantic alignment process, segmentation networks struggle to manage the degradation of discriminative features caused by the haze effect. This results in inaccurate semantic segmentation outcomes, which, in turn, adversely affects the subsequent semantic reconstruction process. For example, areas of the sky might be incorrectly identified as bodies of water due to the reduced clarity and contrast typical of hazy images. On the contrary, the proposed intrinsic semantic alignment leverages the inherent consistency in semantic information between the visible and near-infrared spectra. By exploiting this intrinsic correspondence, our HSVF significantly improves the accuracy of semantic segmentation results. This enhancement is pivotal, as more precise segmentation lays a solid foundation for improved reconstruction processes and later haze removal.

\noindent\textbf{Influence of joint attention mechanism.}
Since we primary focus on the visual structure preservation in the proposed visual appearance fusion module, we further analyze the influence of different attention mechanisms. We utilize widely-used image fusion metrics MI, VIF and $Q_{AB/F}$ for evaluation. As shown in Tab. \ref{tab3}, the proposed self- and cross-attention mechanism achieves better information preservation compared with either intra-modal self-attention or inter-modal cross-attention, further validating the effectiveness of the proposed joint self- and cross-attention scheme.

\subsection{Analysis and Discussion}
\noindent\textbf{Analysis of semantic guidance level for reconstruction.}
We conducted a detailed analysis of the influence of different semantic guidance levels to scene reconstruction. As shown in Fig. \ref{fig_analysis_keycomponents}, we compared the results with different settings: without semantic guidance, with image-level semantic guidance, and with region-level semantic guidance. The analysis reveals that directly employing GANs to reconstruct clear scenes without semantic understanding fails to preserve the scene's content integrity. For example, building areas might be reconstructed with vegetation textures. This occurs due to the lack of semantic understanding impacting the network's ability to maintain the original scene's content. The incorporation of image-level semantic guidance improves the network's capability to preserve the intrinsic content of the scene during reconstruction, reducing the likelihood of such discrepancies. However, the results may still exhibit artifacts due to the implicit guidance. By implementing region-level semantic guidance, the network can reconstruct scenes that are not only more realistic but also maintain semantic consistency. This approach significantly enhances the fidelity of the reconstructed scenes, particularly in maintaining the correct attributes and features for different regions.

\begin{figure}[t]
	\centering
	\includegraphics[width=\linewidth]{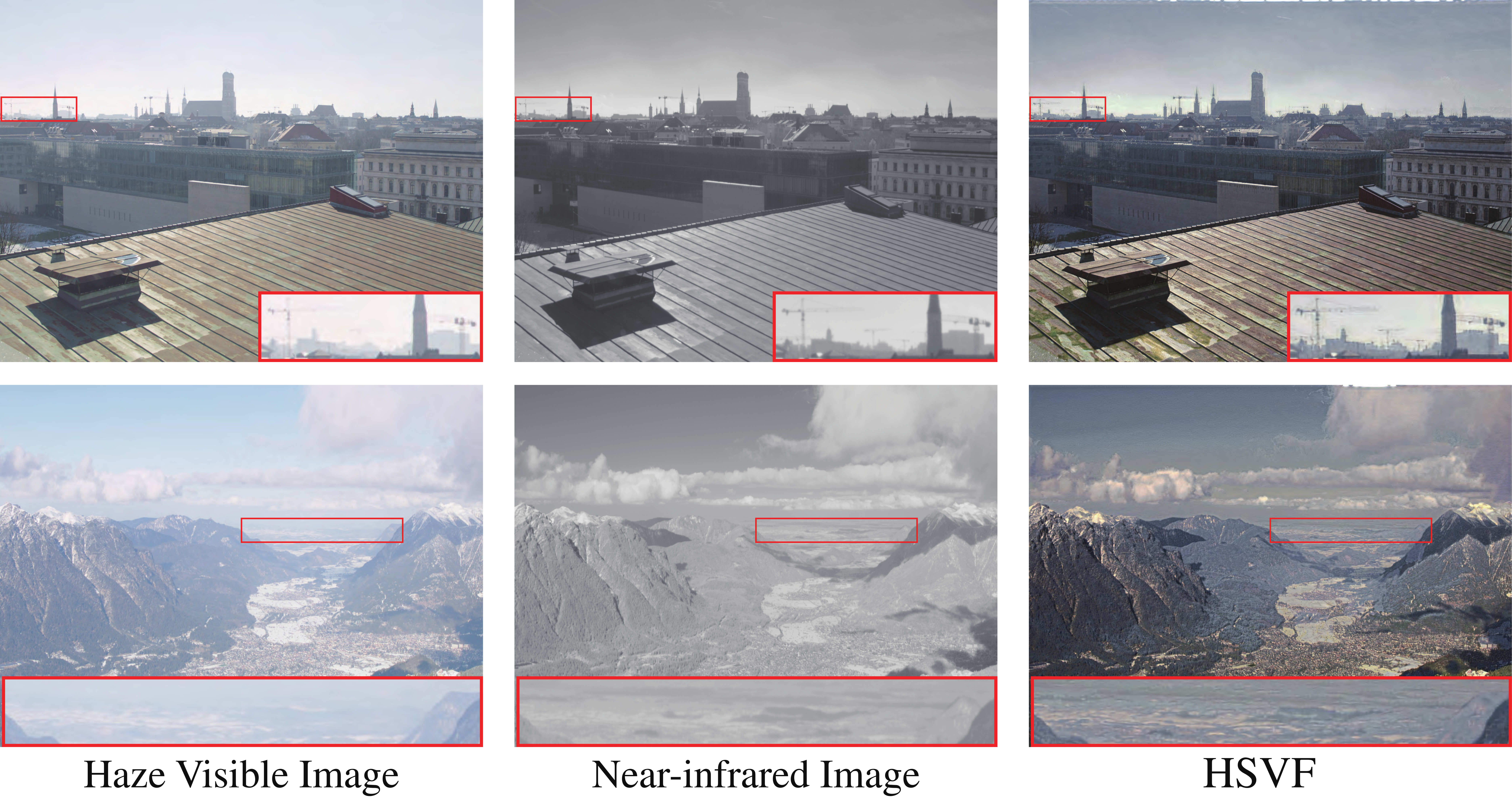}
	\caption{Analysis of the generalizability on ARRI RGB/NIR dataset \cite{luthen2017rgb}.}
	\label{fig_analysis_generalization}
\end{figure}

\noindent\textbf{Analysis of different semantic-visual loss weight.}
We conducted a detailed hyper-parameter analysis to investigate how varying the relative weights between the reconstruction and fusion losses affects performance. Specifically, we evaluated different weight ratios and two extreme cases in which either the reconstruction or fusion loss was entirely removed. Visual comparisons are shown in Fig. \ref{fig_analysis_parameter}, and quantitative results in terms of FADE and NIQE are summarized in Tab. \ref{tab_parameter_analysis}. From the visual results, when the reconstruction loss dominates, the outputs better reconstruct a clear scene but lose fine details; conversely, when the fusion loss dominates, multimodal details are better preserved but residual haze remains, reducing contrast. Quantitatively, the NIQE score worsens as the reconstruction loss weight increases, indicating a degradation of high-frequency details. Meanwhile, the FADE score first improves, then declines as reconstruction loss becomes excessively large, suggesting that while moderate reconstruction enhances contrast, excessive weight of reconstruction leads to structural degradation. These findings highlight the importance of balancing clear scene reconstruction and fine texture preservation for optimal long-range haze removal, and empirically justify the chosen hyper-parameters in HSVF.

\noindent\textbf{Analysis of different imaging ranges.}
We further analyze the haze removal performance across different scene depths. In Fig. \ref{fig_analysis_imagingdistance}, we visualize the haze removal results of the proposed HSVF under different ranges of scene depth, varying from less than 1 km to over 10 km. We observe that the proposed HSVF achieves outstanding performance in both semantic segmentation and long-range haze removal across datasets with varying depths of field, which successfully retrieves the semantic scene content (e.g. distance mountain) and visual structure (e.g. window texture of distant building).

\noindent\textbf{Analysis of modalities for long-range scene segmentation.} We further illustrate the importance of multimodal input for accurate semantic segmentation. As shown in Fig. \ref{fig_analysis_modalitysemantic}, we visualize segmentation results with and without haze-robust near-infrared information, in which we adopt HRNet \cite{wang2020deep} for semantic segmentation based solely on visible images. We can observe that without the guidance of near-infrared, the results of segmentation may struggle to distinguish the boundary of distant scene content (e.g. the distant buildings and tower cranes). On the contrary, thanks to the guidance provided by the near-infrared and intrinsic semantic alignment module, the proposed HSVF could achieve better segmentation results of distant scene content, so as to better reconstruct haze-free scene content and facilitate long-range imaging results.

\begin{figure}[t]
	\centering
	\includegraphics[width=\linewidth]{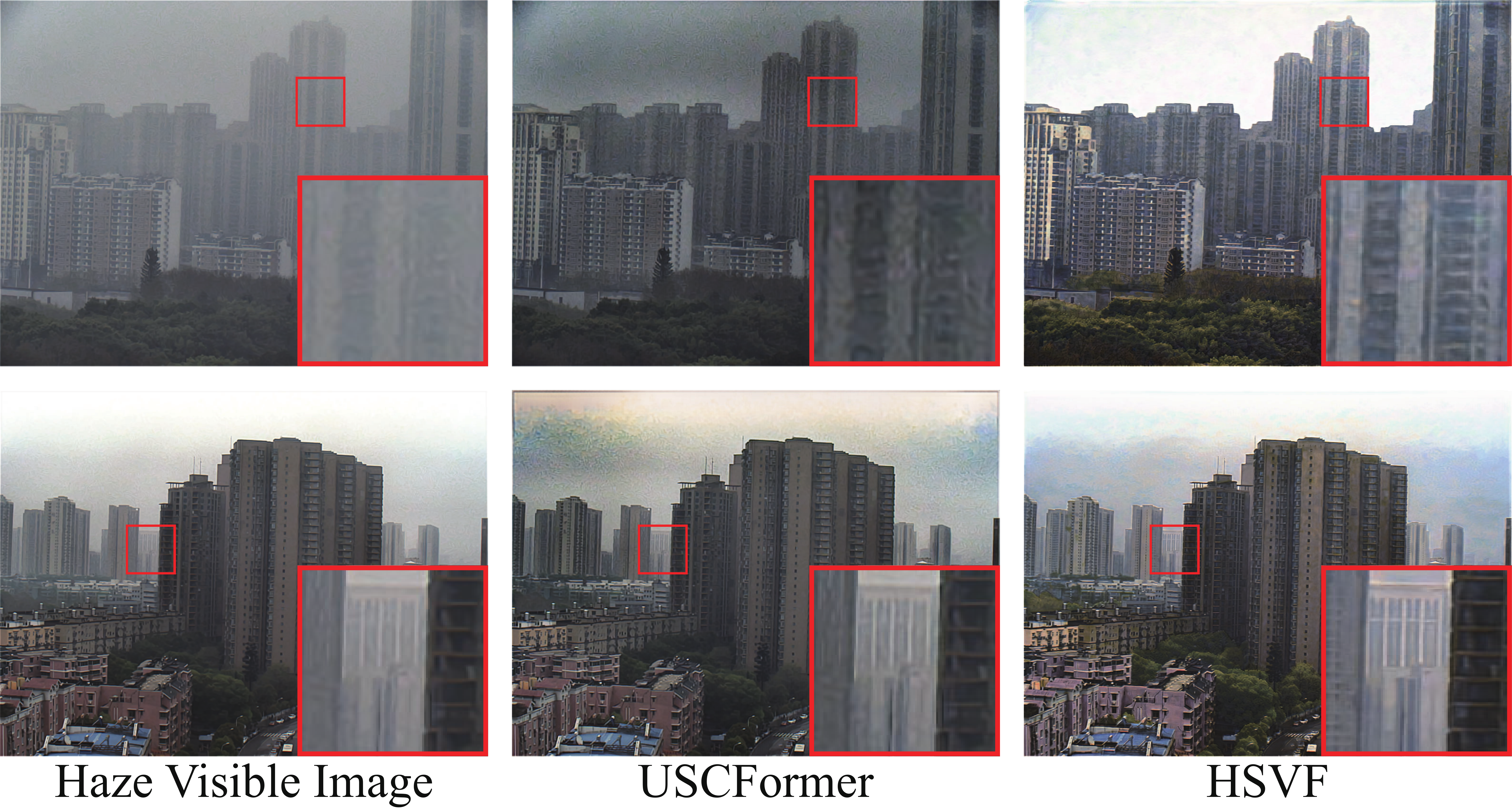}
	\caption{Analysis of the semantic utilization strategy of the proposed HSVF and other semantic-guided dehazing methods USCFormer \cite{wang2023uscformer}.}
	\label{fig_analysis_semanticutilization}
\end{figure}

\noindent\textbf{Generalization of HSVF to other datasets.} 
We then conducted further analysis of the generalizability of our proposed method on datasets beyond our primary VNHD dataset. Specifically, we applied the weights trained on VNHD to the ARRI RGB/NIR dataset \cite{luthen2017rgb} without any additional fine-tuning or domain-specific adaptations. The evaluation results, as shown in Fig. \ref{fig_analysis_generalization}, demonstrate that our method performs consistently well in removing haze even in challenging long-range scenarios captured by the ARRI RGB/NIR dataset without fineturning process. This outcome highlights the robustness of the proposed approach in handling diverse imaging conditions and spectral modalities.

\noindent\textbf{Analysis of semantic utilization strategy for dehazing.}
We further analyze the semantic utilization strategy for dehazing by comparing long-range haze removal results of the proposed HSVF with USCFormer \cite{wang2023uscformer}, which focuses on semantic feature fusion to facilitate dehazing. As shown in Fig. \ref{fig_analysis_semanticutilization}, we can observe that USCFormer, which utilizes image-level semantic feature fusion to directly guide dehazing process, could indeed enhance the contrast of the scene and restore the sky region. However, they may face difficulties in restoring the details of distant scenes. On the contrary, the proposed HSVF not only adopts region-level semantics to explicitly guide scene reconstruction for improved scene clarity, but also utilizes visual fusion for retrieving the lost scene details, resulting in more satisfying long-range haze removal performance.

\begin{table}[t] \centering
	\caption{Running time comparison. We compare HSVF with supervised dehazing method Dehamer \cite{guo2022image}, C2PNet \cite{zheng2023curricular}, DA-Clip \cite{luo2024controlling}, AdaIR \cite{cui2025adair}; semi- or unsupervised dehazing methods PSD \cite{chen2021psd}, RIDCP \cite{wu2023ridcp}, UME-Net \cite{sun2024unsupervised}, D4+ \cite{yang2024robust}; and fusion methods SwinFusion \cite{ma2022swinfusion}, MHLP \cite{yang2023detail}, CDDFuse \cite{zhao2023cddfuse}, EMMA \cite{zhao2024equivariant}.}
	\label{tab_running_time}
	\setlength{\tabcolsep}{5mm}
	\begin{tabular}{c|l|c}
		\toprule[1.5pt]
		Type & Methods     & Time(s) \\
		\midrule[1.5pt]
		\multirow{4}{*}{\shortstack{Supervised}}
		&Dehamer      & 0.63  \\
		&C2PNet       & 1.86  \\
		&DA-Clip      & 70.83  \\
		&AdaIR        & 4.43  \\\midrule[0.5pt]
		\multirow{4}{*}{\shortstack{Semi- or\\Un-supervised}}
		&PSD          & 1.12  \\
		&RIDCP        & 0.97  \\
		&UME-Net      & 1.56  \\
		&D4+          & 0.52  \\\midrule[0.5pt]
		\multirow{4}{*}{\shortstack{Fusion-based}}
		&SwinFusion   & 3.54  \\
		&MHLP         & 10.69  \\
		&CDDFuse      & 0.97  \\
		&EMMA         & 0.22  \\\midrule[0.5pt]
		--- &HSVF     & 3.30\\
		\bottomrule[1.5pt]
	\end{tabular}
\end{table}

\noindent\textbf{Limitation of HSVF.}
Table \ref{tab_running_time} shows the results of runtime comparison between our proposed HSVF method and other representative state-of-the-art methods. The comparison reports the average inference time (in seconds) per 1024$\times$1024 image on a single NVIDIA 3090 GPU (deep learning methods) or an Intel(R) Core(TM) i9-10900K CPU@ 3.70GHz (conventional methods). The results from the table reflect a current limitation of our work: the primary focus has been on improving restoration quality rather than optimizing inference speed. In our future work, we plan to explore strategies for enhancing the inference efficiency of HSVF and developing practical real-time solutions for real-world deployment.

\section{Conclusion}
\label{sec:conclusion}
In this work, we tackle the under-explored problem of long-range haze removal by introducing a Hierarchical Semantic-Visual Fusion (HSVF) framework. HSVF jointly leverages high-level semantic consistency and low-level multimodal structural cues to address severe signal degradation and detail loss caused by intensified scattering over long distances. Specifically, the semantic stream learns modality-invariant representations to achieve haze-robust semantic prediction. Then the acquired semantics serves as strong priors for reconstructing high-contrast, clear scenes. In parallel, the visual stream complements missing texture details by fusing structural information from near-infrared and visible images through joint self- and cross-attention mechanisms. By integrating the two streams, HSVF restores both global contrast and rich details under heavy haze. Additionally, we introduce a pixel-aligned visible-infrared haze dataset with semantic annotations as a valuable benchmark for long-range haze removal. Extensive experiments demonstrate the superiority of HSVF over state-of-the-art methods, highlighting its potential for long-range imaging applications such as video surveillance \cite{cheng2022hybrid, cheng2025semantic, he2025exploring}.

\bibliographystyle{IEEEtran}
\bibliography{main.bib}

\begin{IEEEbiography}[{\includegraphics[width=1in,height=1.25in,clip,keepaspectratio]{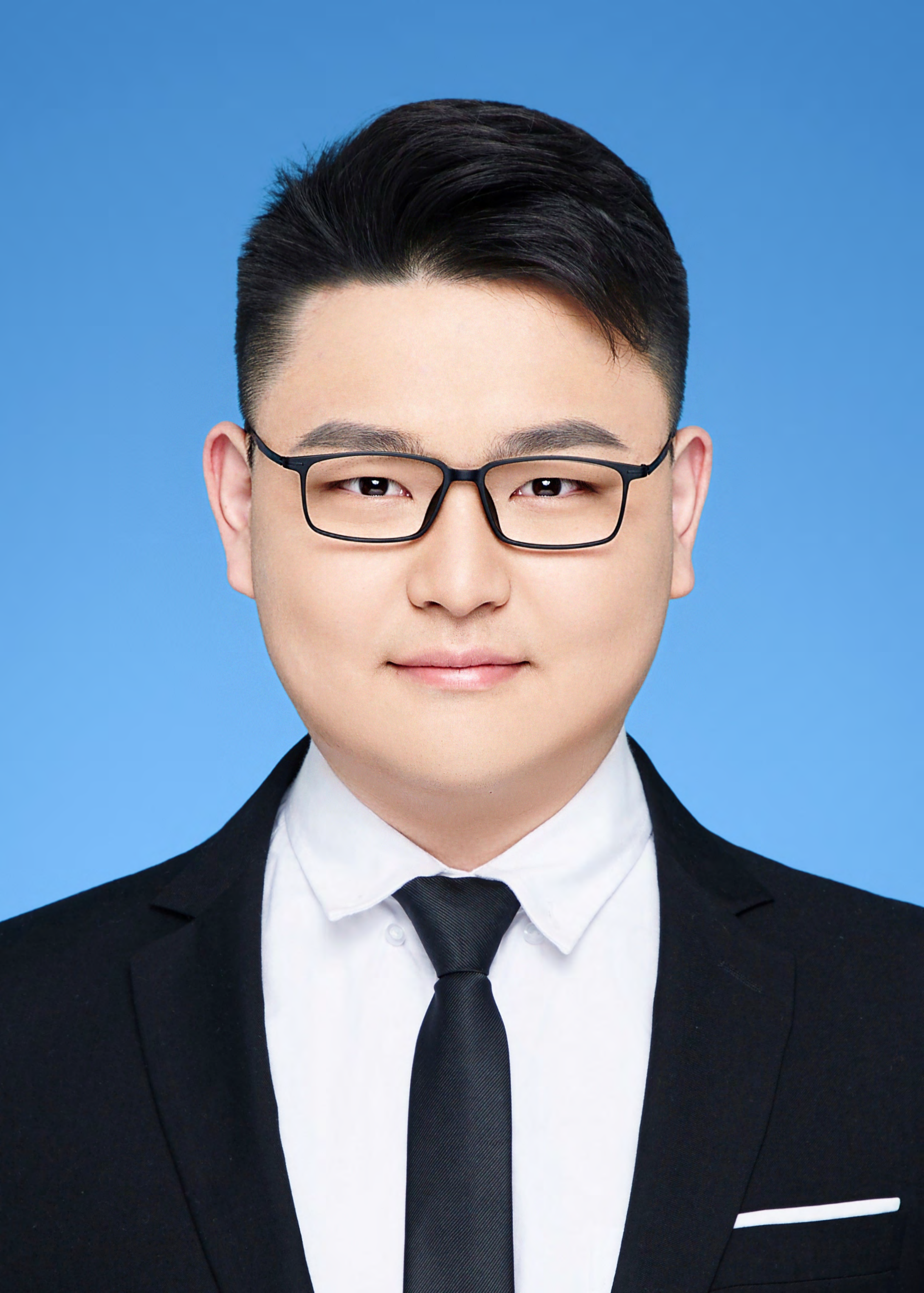}}]Yi Li received B.S. degree in School of Artificial Intelligence and Automation from Huazhong University of Science and Technology, Wuhan, China in 2018. He is currently presuming Ph.D. degree in School of Artificial Intelligence and Automation, HUST. His research interests include image restoration and scene understanding under adverse weather.
\end{IEEEbiography}

\begin{IEEEbiography}[{\includegraphics[width=1in,height=1.25in,clip,keepaspectratio]{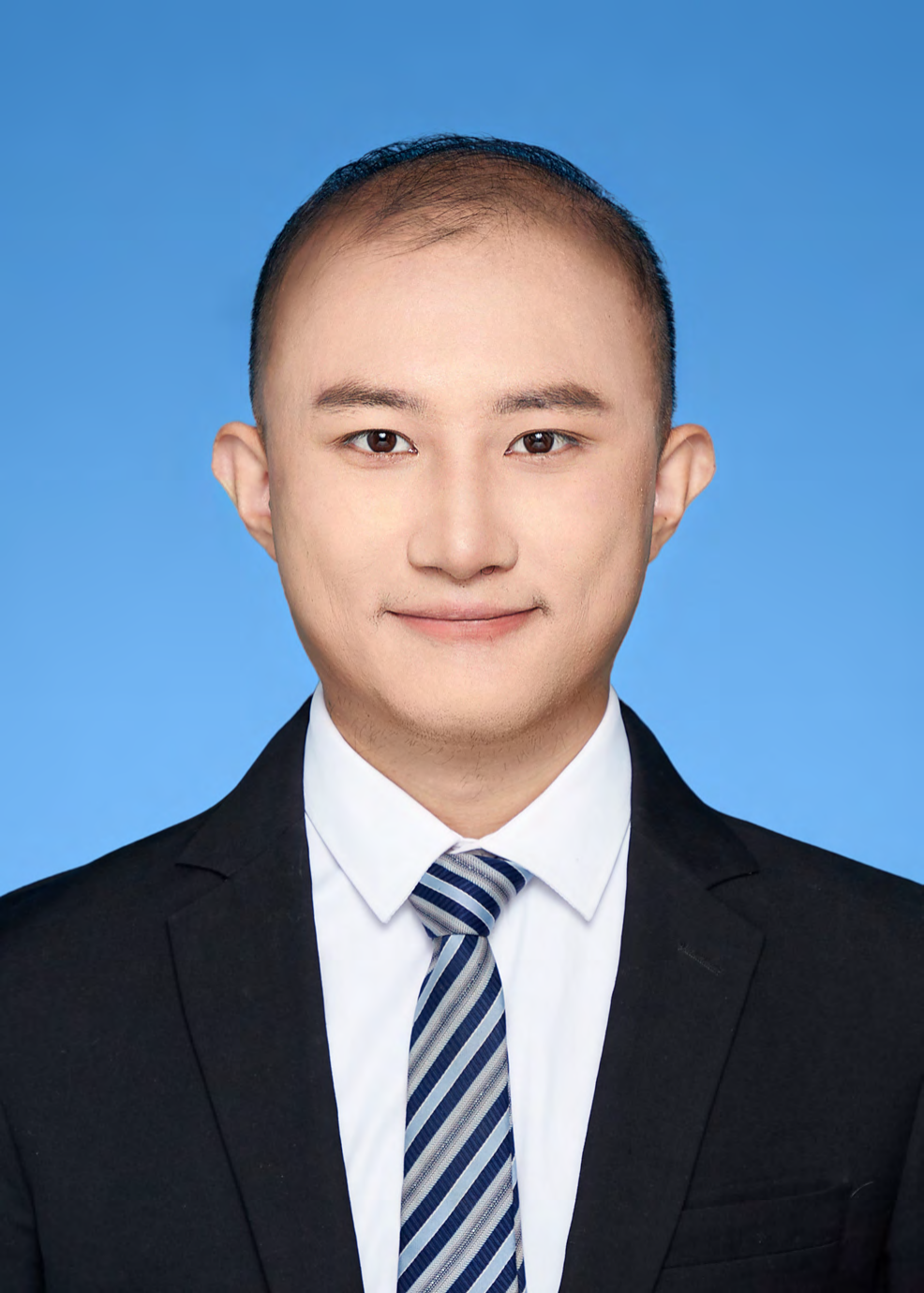}}]{Xiaoxiong Wang} received the B.S. degree in College of Information Science and Engineering from Northeastern University, Shenyang, China, in 2023. Currently, he is pursuing the M.S. degree with the School of Artificial Intelligence and Automation, Huazhong University of Science and Technology, Wuhan, China. His interests include image restoration and space target keypoint detection.
\end{IEEEbiography}

\begin{IEEEbiography}[{\includegraphics[width=1in,height=1.25in,clip,keepaspectratio]{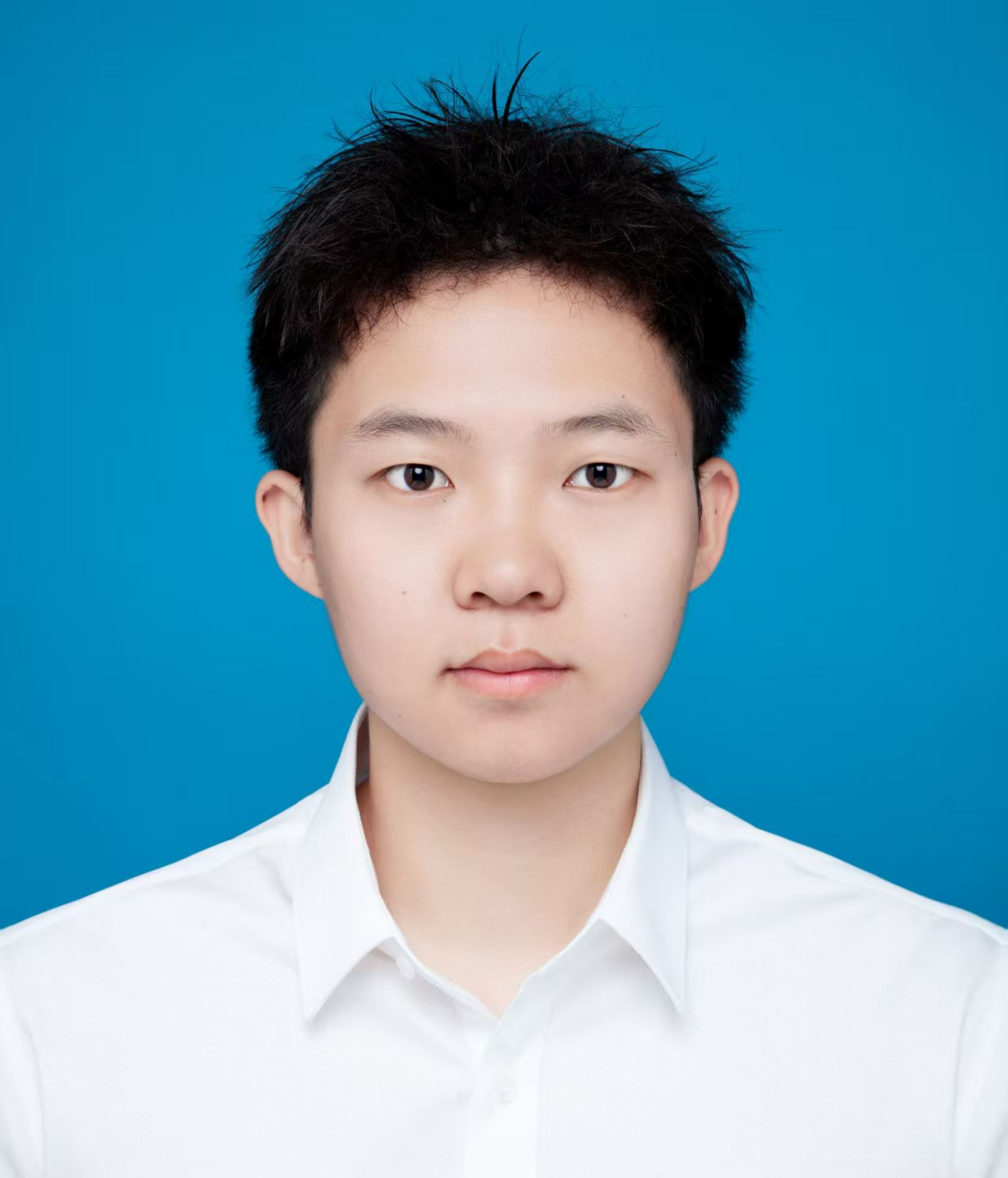}}]{Jiawei Wang} received his M.S. degree from Huazhong University of Science and Technology, Wuhan, China in 2024. His research interests include multi–spectral sementic segmentation in low–visibility scenarios
\end{IEEEbiography}

\begin{IEEEbiography}[{\includegraphics[width=1in,height=1.25in,clip,keepaspectratio]{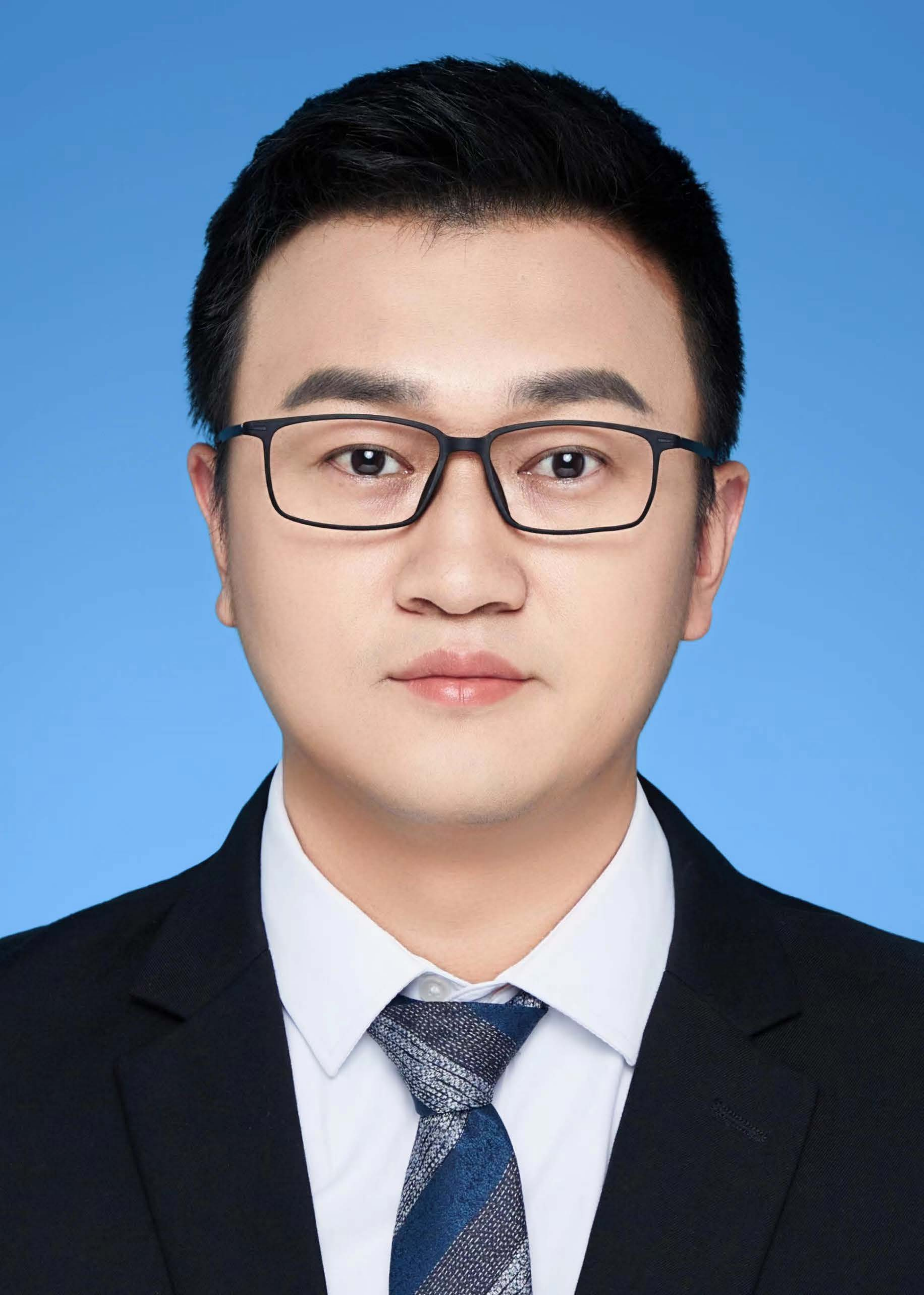}}]{Yi Chang} received B.S. degree from University	of Electronic Science and Technology of China,	Chengdu, China, in 2011, and M.S. degree and Ph.D. from Huazhong University of Science and Technology (HUST), in 2014 and 2019, respectively. Currently, he is a Huazhong distinguished young scholar with School of Artificial Intelligence and Automation, HUST. His research interests include adverse weather image restoration and understanding, neuromorphic vision, and multispectral image processing.
\end{IEEEbiography}

\begin{IEEEbiography}[{\includegraphics[width=1in,height=1.25in,clip,keepaspectratio]{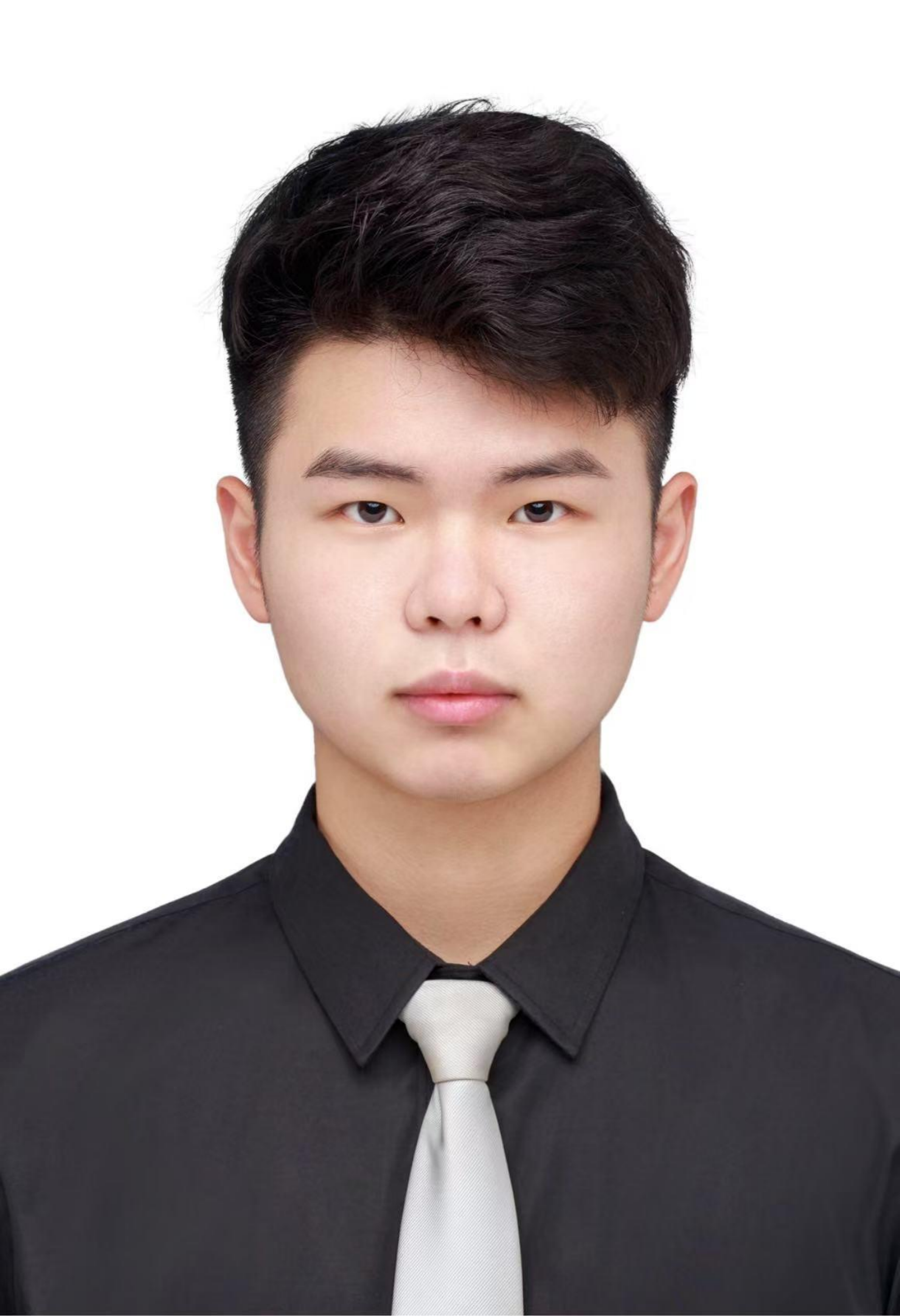}}]{Kai Cao} is currently presuming B.S. degree in School of Computer Science and Technology, Huazhong University of Science and Technology, Wuhan, China. His research interests include semantic segmentation and scene parsing under adverse weather.
\end{IEEEbiography}

\begin{IEEEbiography}[{\includegraphics[width=1in,height=1.25in,clip,keepaspectratio]{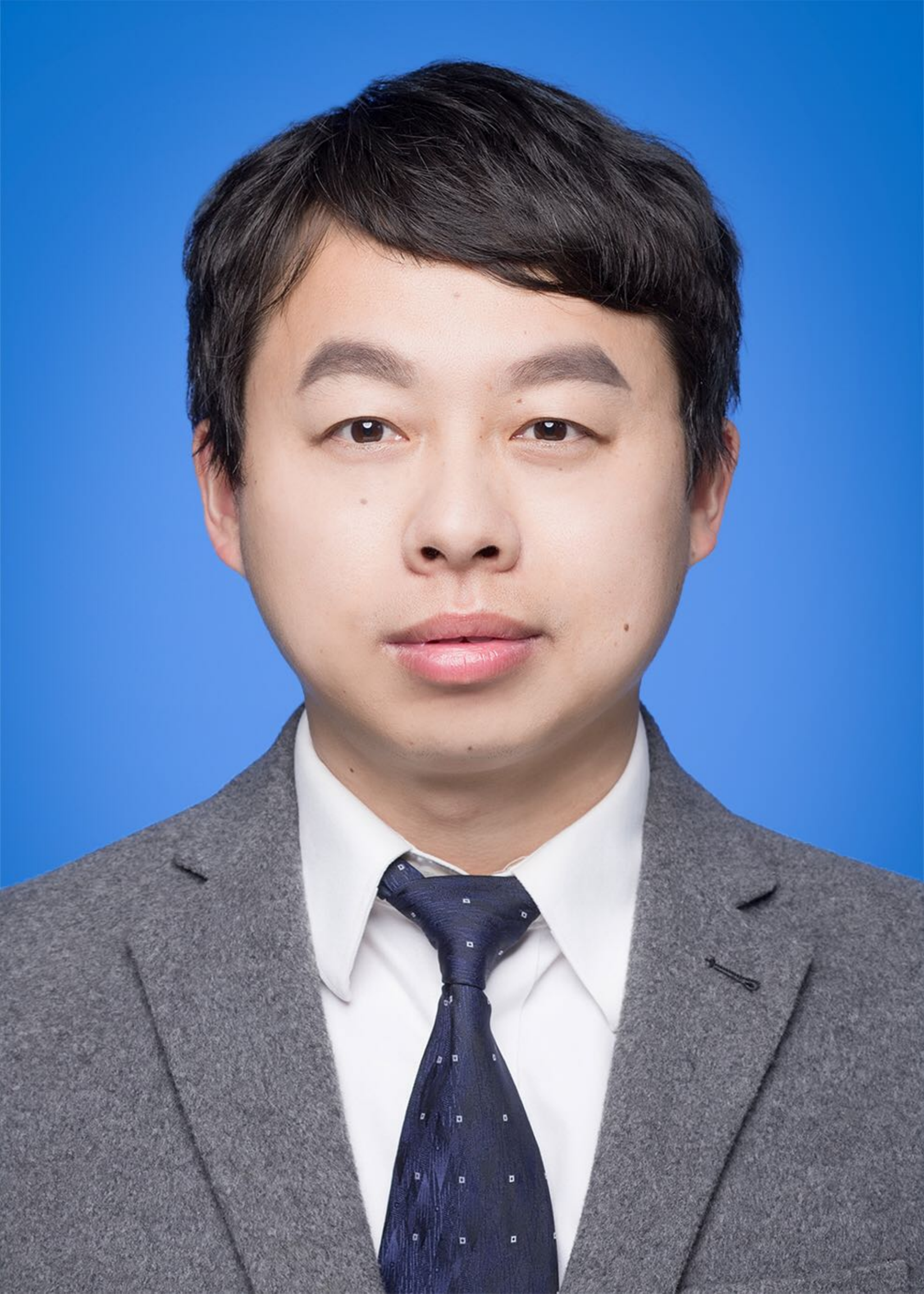}}]{Luxin Yan} received the B.S. and the Ph.D. degree from Huazhong University of Science and Technology (HUST) in 2001 and 2007, respectively. He is currently a Huazhong distinguished chief professor with School of Artificial Intelligence and Automation, HUST, and is also the deputy director of National Key Laboratory of Multispectral Information Intelligent Processing Technology. His research interests include multispectral image processing, pattern recognition and real-time embedded system.
\end{IEEEbiography}

\vfill

\end{document}